\documentclass[journal=jacsat,manuscript=article]{achemso}

\usepackage[version=3]{mhchem} 
\usepackage{multirow}
\usepackage{makecell}
\usepackage{cellspace}
\usepackage{booktabs}
\usepackage{amsmath} 
\usepackage{amssymb} 
\usepackage{enumitem}
\usepackage{algorithm}
\usepackage{algpseudocode}
\usepackage{lipsum}

\author{Janghoon Ock}
\affiliation{Department of Chemical Engineering, Carnegie Mellon University, 5000 Forbes Avenue, Pittsburgh, PA 15213, USA}
\alsoaffiliation[unl-cheme]
{Department of Chemical and Biomolecular Engineering, University of Nebraska--Lincoln, Lincoln, NE 68588, USA}
\author{Radheesh Sharma Meda}
\affiliation{Department of Chemical Engineering, Carnegie Mellon University, 5000 Forbes Avenue, Pittsburgh, PA 15213, USA}
\author{Srivathsan Badrinarayanan}
\affiliation
{Department of Chemical Engineering, Carnegie Mellon University, 5000 Forbes Avenue, Pittsburgh, PA 15213, USA}
\author{Neha S Aluru}
\affiliation{School of Engineering Medicine, Texas A\&M University, Houston, TX 77030, USA}
\author{Achuth Chandrasekhar}
\affiliation{Department of Material Science and Engineering, Carnegie Mellon University, 5000 Forbes Avenue, Pittsburgh, PA 15213, USA}
\author{Amir Barati Farimani}
\affiliation{Department of Mechanical Engineering, Carnegie Mellon University, 5000 Forbes Avenue, Pittsburgh, PA 15213, USA}
\email{barati@cmu.edu}

\title[An \textsf{achemso} demo]
  {Large Language Model Agent for Modular Task Execution in Drug Discovery}

\abbreviations{IR,NMR,UV}

\begin{document}


\begin{abstract}
We present a modular framework powered by large language models (LLMs) that automates and streamlines key tasks across the early-stage computational drug discovery pipeline. By combining LLM reasoning with domain-specific tools, the framework performs biomedical data retrieval, literature-grounded question answering via retrieval-augmented generation, molecular generation, multi-property prediction, property-aware molecular refinement, and 3D protein–ligand structure generation. The agent autonomously retrieved relevant biomolecular information, including FASTA sequences, SMILES representations, and literature, and answered mechanistic questions with improved contextual accuracy compared to standard LLMs. It then generated chemically diverse seed molecules and predicted 75 properties, including ADMET-related and general physicochemical descriptors, which guided iterative molecular refinement. Across two refinement rounds, the number of molecules with QED $>$ 0.6 increased from 34 to 55. The number of molecules satisfying empirical drug-likeness filters also rose; for example, compliance with the Ghose filter increased from 32 to 55 within a pool of 100 molecules. The framework also employed Boltz-2 to generate 3D protein–ligand complexes and provide rapid binding affinity estimates for candidate compounds. These results demonstrate that the approach effectively supports molecular screening, prioritization, and structure evaluation. Its modular design enables flexible integration of evolving tools and models, providing a scalable foundation for AI-assisted therapeutic discovery.
\end{abstract}

\section{Introduction}

The discovery of drug-like molecules is fundamental to therapeutic innovation and the advancement of public health. However, identifying viable candidates is a highly complex and resource-intensive process. It requires navigating an enormous chemical space, integrating heterogeneous biological and chemical data sources, and conducting iterative experimental validation. Empirically, drug development spans 10-15 years and can cost \$1.5 - 2.5 billion per approved therapy. Much of this burden arises from attrition in Phase II and III clinical trials, where insufficient efficacy or safety issues often lead to late-stage failures~\cite{Paul2010-sr, Roland2024-pw, Subbiah2023-tw}. A key strategy to reduce this attrition is to identify high-quality candidates as early as possible, thereby improving the likelihood of clinical success. At the same time, the chemical search space is astronomically large, with estimates of more than $10^{60}$ drug-like molecules, making it impossible to explore effectively with traditional trial-and-error approaches~\cite{Bohacek1996, Polishchuk2013}.

To mitigate these challenges, computational approaches have become essential in early-stage drug development. Methods such as molecular docking \cite{Fan2019, autodock_vina, docking_software}, quantitative structure–activity relationship (QSAR) modeling \cite{qsar, qsar_2, dl_qsar}, and molecular dynamics simulations \cite{md_for_protein_1, md_for_protein_2, gradnav} have significantly improved the speed and accuracy of compound screening and optimization. More recently, machine learning (ML) techniques have expanded this toolkit by enabling predictive modeling of pharmacokinetic and pharmacodynamic properties, de novo molecular generation, and multitask learning across diverse biomedical tasks\cite{badrinarayanan2025mofgptgenerativedesignmetalorganic, peptidebert, idpbert, multipeptide}. AlphaFold has achieved high-accuracy protein structure prediction~\cite{alphafold, alphafold2, alphafold3}.  In parallel, generative models such as Variational Autoencoders (VAEs), Generative Adversarial Networks (GANs), and diffusion models (e.g., JT-VAE, MolGAN, MolDiff) enable de novo or target-specific molecular design~\cite{jt_vae, molgan, moldiff}. Also, Graph Neural Networks (GNNs) and transformer-based models improve property prediction~\cite{admetlab2, moleculenet, molprop, deeptox}. Collectively, these methods increase the scalability and automation potential of modern drug discovery workflows.

Despite this progress, most computational methods remain task-specific and require manual orchestration by domain experts. Drug discovery is inherently a multi-step, interdependent process that requires seamless integration across diverse tasks \cite{Berdigaliyev2020, Hughes2011, SINHA201819}. This fragmented implementation limits the scalability and efficiency of current pipelines. As the demand for faster and more cost-effective drug screening grows, there is an urgent need for unified, intelligent platforms capable of autonomously coordinating these tasks while supporting expert decision-making.

Large Language Models (LLMs) offer a compelling solution to this need. Trained on massive corpora of natural language data, LLMs exhibit strong reasoning capabilities and domain-agnostic knowledge. When augmented with external tools, such as domain-specific plugins, APIs, and software libraries, LLM-based agents can overcome the limitations of general-purpose language models and act as interpretable, flexible controllers of scientific workflows \cite{aiscientist, coscientist}. 

LLM agents have recently been applied to automate diverse aspects of scientific discovery, including experimental design, material synthesis planning, and data analysis \cite{chemcrow, adsorb_agent, llm_design, llm3dprint, matsci_agent}. For instance, the dZiner framework leverages LLM agents for molecular design through iterative reasoning and structure-based optimization\cite{dziner}. CACTUS\cite{McNaughton2024-bp} demonstrated the use of an LLM to wrap cheminformatics utilities for evaluating properties such as Mwt, LogP, TPSA, and drug-likeness filters, providing a practical but narrowly scoped toolset. In contrast, SciToolAgent\cite{Ding2025-he} introduced a knowledge-graph–driven orchestration layer that allows an LLM to select from hundreds of scientific tools across domains ranging from biological to materials, illustrating scalability but lacking a dedicated workflow for therapeutic design. Building on these directions, DrugPilot\cite{Li2025-iy} focuses on orchestrating multi-stage discovery workflows using parameterized reasoning, primarily demonstrated on established model-zoo tasks and benchmark datasets rather than end-to-end molecular design. A more detailed overview of the comparative landscape of agentic frameworks is provided in Supplementary Information Section 9.


We introduce AgentD, an LLM-powered agent framework designed to support and streamline the drug discovery pipeline. The agent performs a wide range of important tasks, including biomedical data retrieval from structured databases and unstructured web sources, answering domain-specific scientific queries, generating seed molecule libraries via SMILES-based generative models, predicting a broad spectrum of drug-relevant properties, refining molecular representations to improve drug-likeness, and generating 3D molecular structures for downstream analysis. Our results demonstrate that this agent-driven framework streamlines early-phase drug discovery and serves as a flexible foundation for scalable, AI-assisted therapeutic development. Its modular architecture allows for continual improvement as more advanced tools and models become available.

\section{Agentic Workflow}
\subsection{Task Modules}

\begin{figure*}[!htb] 
\centering
\includegraphics[width=0.99\textwidth]{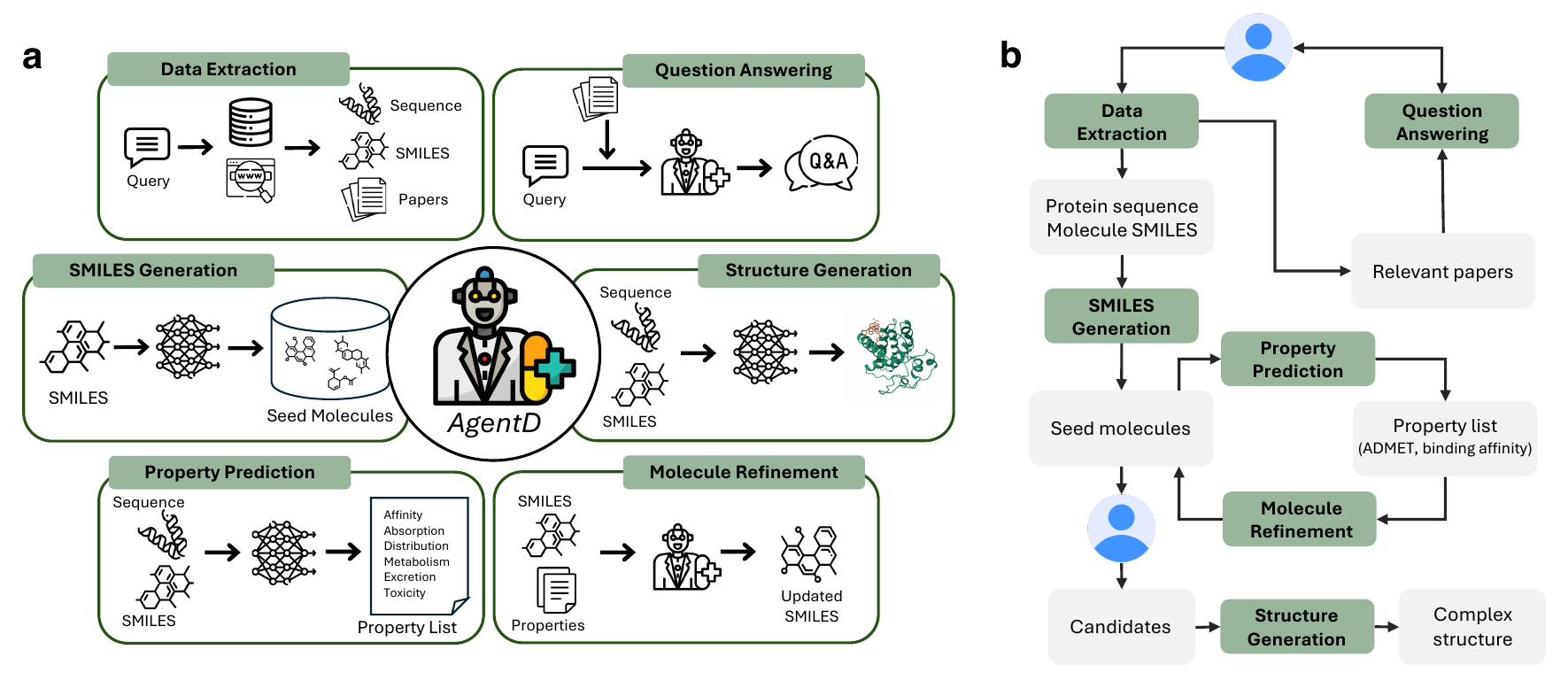} 
\caption{AgentD overview. \textbf{a} Task modules supported by AgentD. In question-answering and molecular-refinement tasks, outputs are generated directly by the language model, while other tasks rely on external computational tools for final results. \textbf{b} Workflow illustrating how each task module contributes to the overall drug discovery pipeline.}
\label{fig:overview}
\end{figure*}

AgentD performs six essential tasks across the drug discovery pipeline, as illustrated in Figure~\ref{fig:overview}a. From a code-design perspective, the agent integrates LLMs sourced from OpenAI, Anthropic, and DeepSeek. These models function both as the core reasoning module and as the natural-language interaction layer. In the present implementation, AgentD is primarily developed and executed using OpenAI’s GPT-4o as the main language model. By integrating domain-specific tools and databases, AgentD coordinates a wide range of activities - from data retrieval and molecular generation to property evaluation and structure prediction. The primary tool components supporting each task module are summarized in Table~\ref{tab:agent-tools}.

\textbf{Data Extraction.} AgentD is capable of retrieving biomedical information from both structured and unstructured sources. Given a query, such as identifying known drugs associated with a specific protein–disease context, the agent retrieves the protein’s FASTA sequence from UniProt~\cite {uniprot}, searches for relevant drug names on the web, and extracts corresponding SMILES representations from ChEMBL~\cite{chembl}. Additionally, AgentD autonomously constructs keyword-based queries to download relevant open-access scientific literature from Semantic Scholar, providing a broader context for downstream reasoning and decision-making.

\textbf{Question Answering.} In therapeutic applications, high accuracy and mechanistic specificity are critical. Generic responses that only appear plausible are inadequate for addressing domain-specific questions relevant to drug discovery and biomedical research. To handle domain-specific scientific queries, AgentD employs a retrieval-augmented generation (RAG) strategy. By grounding its responses in literature obtained during the data extraction phase, the agent provides context-aware and evidence-based answers.

\textbf{SMILES Generation.} Constructing a diverse and chemically relevant seed molecule pool is essential for effective early-stage virtual screening. AgentD generates seed molecules in SMILES format using external generative models. In this study, we incorporate two such models: REINVENT~\cite{reinvent2, reinvent4}, which supports de novo molecule generation without requiring input SMILES, and Mol2Mol~\cite{mol2mol}, which performs conditional generation to produce molecules structurally similar to a given input SMILES. This dual capability enables both exploration and exploitation in the molecular search space.

\textbf{Property Prediction.} For each candidate molecule, AgentD predicts key pharmacologically relevant properties, including ADMET (absorption, distribution, metabolism, excretion, and toxicity) profiles and binding affinity (e.g., $pK_d$). ADMET prediction is performed using the Deep-PK API~\cite{deep_pk}, which accepts SMILES strings as input. For binding affinity estimation, the BAPULM model~\cite{bapulm} is used, which operates on both the SMILES and the protein’s FASTA sequence. These predictions help prioritize compounds based on both efficacy and safety.

\textbf{Molecule Refinement.} Based on the predicted properties, AgentD can identify molecular shortcomings such as toxicity or poor permeability, and propose targeted structural modifications to improve attributes like solubility and metabolic stability. This SMILES refinement is carried out solely through the LLM’s internal reasoning and built-in chemical knowledge, without additional model-based tools \cite{dziner}. This task is carried out on top of the property prediction stage.

\textbf{Structure Generation.} AgentD can generate 3D structures of protein–ligand complexes using Boltz-2 as an external tool~\cite{wohlwend2024boltz1, passaro2025boltz2}. This process produces candidate complex structures along with associated binding metrics such as IC\textsubscript{50} values and inhibitor probability. These structures can serve as inputs for downstream computational tasks such as docking simulations or molecular dynamics, offering deeper insight into the biophysical interactions of the drug candidates.

\begin{table}[!hbtp]
\centering
\caption{Key external tools integrated into each task module within the AgentD drug discovery pipeline.}
\label{tab:agent-tools}
\resizebox{\textwidth}{!}{%
\begin{tabular}{ll}
\toprule
\textbf{Task Module} & \textbf{Primary Tool Components} \\
\midrule
Data Extraction & UniProt~\cite{uniprot}, ChEMBL~\cite{chembl}, Serper~\cite{serperAPI}, Semantic Scholar APIs \\
Question Answering & RAG, LLM internal knowledge and reasoning \\
SMILES Generation & REINVENT~\cite{reinvent4}, Mol2Mol~\cite{mol2mol} \\
Property Prediction & DeepPK API~\cite{deep_pk}, BAPULM~\cite{bapulm}  \\
Molecule Refinement & LLM internal knowledge and reasoning\textsuperscript{*} \\
Structure Generation & Boltz-2~\cite{passaro2025boltz2} \\
\bottomrule
\end{tabular}
}
{\raggedleft
\textsuperscript{*}\footnotesize Adapted from the dZiner framework~\cite{dziner}.\par}
\end{table}

\subsection{Workflow}

All six task modules in AgentD are designed to support key components of the drug discovery pipeline, as illustrated in Figure~\ref{fig:overview}b. The process begins with a user-provided query, such as identifying drug molecules that target a specific disease-related protein. Through the data extraction module, AgentD retrieves the protein’s FASTA sequence from UniProt and identifies known drugs using sources such as ChEMBL and Google search APIs. In this study, we demonstrate the workflow using three representative therapeutic targets: BCL-2 for lymphocytic leukemia, JAK-2 for myelofibrosis, and thrombin for thrombosis. The BCL-2 case is discussed in detail in the main manuscript, while results for the other two targets are provided in Supplementary Information Sections 5 and 6.

The AgentD framework checks whether known drug molecules exist for a given target or disease. If available, these molecules serve as starting points for subsequent molecular exploration. Because AgentD includes a de novo generation module, the workflow can still proceed even when no such molecules are found. The SMILES generation task builds a chemically diverse library of candidate molecules using generative models. These molecules are then evaluated using property prediction tools to estimate key pharmacological characteristics such as ADMET profiles, general physicochemical properties, and binding affinity. Based on these predicted properties, AgentD proposes SMILES-level modifications to improve attributes like solubility, toxicity, and metabolic stability. These refinements rely on the language model’s internal reasoning rather than external optimization tools and help enrich the candidate pool.

Following refinement, domain-specific criteria can be applied by the user to select promising compounds. For these, the structure generation task creates 3D protein–ligand complexes using the ligand SMILES and target protein sequence. This enables more detailed downstream analyses. The structure generation step also outputs auxiliary metrics like predicted IC\textsubscript{50} and inhibitor probability as proxies for binding strength. Although these estimates from Boltz-2 provide useful initial guidance, they should ultimately be validated using more reliable computational methods, such as molecular dynamics simulations. Additional uncertainty information, including Boltz-2–derived affinity estimates, is provided in Supplementary Information Section 10.


Throughout the workflow, users may request clarification or validation of scientific concepts. The question answering module, implemented using RAG, supports this by leveraging literature collected during the data extraction phase. This functionality serves both to enhance the pipeline and to address user-specified scientific queries on demand.

\section{Results}
\subsection{Data Extraction}

\begin{figure*}[!htb] 
\centering
\includegraphics[width=0.99\textwidth]{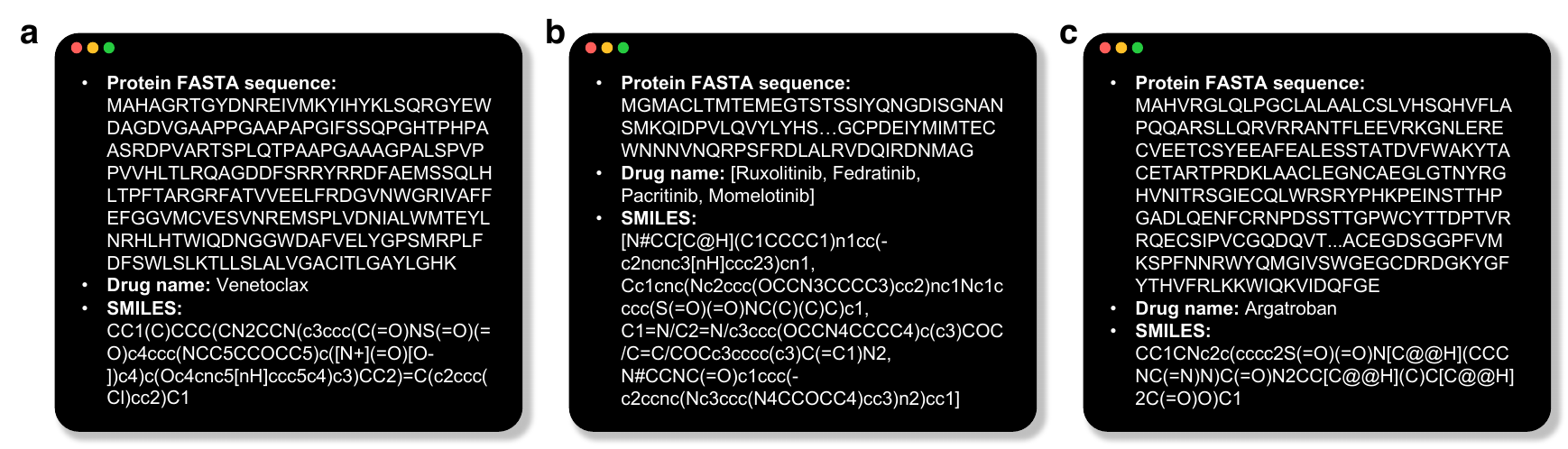} 
\caption{Examples of data extraction. Given a user query specifying a target protein and associated disease, AgentD retrieves the protein's FASTA sequence and identifies known drug molecules relevant to the query. \textbf{a} BCL-2 for lymphocytic leukemia, \textbf{b} JAK-2 for myelofibrosis, and \textbf{c} thrombin for thrombosis.}
\label{fig:data}
\end{figure*}

AgentD integrates web search capabilities and database API access as tool modules for extracting protein and molecule data. When provided with a user query specifying a target protein and associated disease, the agent is instructed to: (i) retrieve the protein’s FASTA sequence, (ii) identify existing drugs relevant to the specified target–disease context, and (iii) download related open-access scientific literature. The agent was tested on example queries including BCL-2 for lymphocytic leukemia, JAK2 for myelofibrosis, and thrombin for thrombosis. The results, summarized in Figure~\ref{fig:data}, show that the agent successfully retrieved key molecular data, such as identifying venetoclax as a BCL-2 inhibitor.


The extracted information is used in several downstream modules. The retrieved SMILES is used to initialize molecule generation models for building a seed library. Both the SMILES and protein sequence serve as inputs to the property prediction and 3D structure generation components. Additionally, the downloaded documents are embedded into a vector database, enabling context-aware retrieval during the agent’s question answering task.

\subsection{Domain-Specific Question Answering}

To address the limitations of generic, unsupported answers in biomedical contexts, AgentD employs RAG to ground its responses in domain-specific literature. During the data extraction phase, the agent formulates keyword-based queries from the target protein and disease, retrieves open-access papers via Semantic Scholar, and stores them in a vector database for later retrieval during question answering.

\begin{figure*}[!htb] 
\centering
\includegraphics[width=0.99\textwidth]{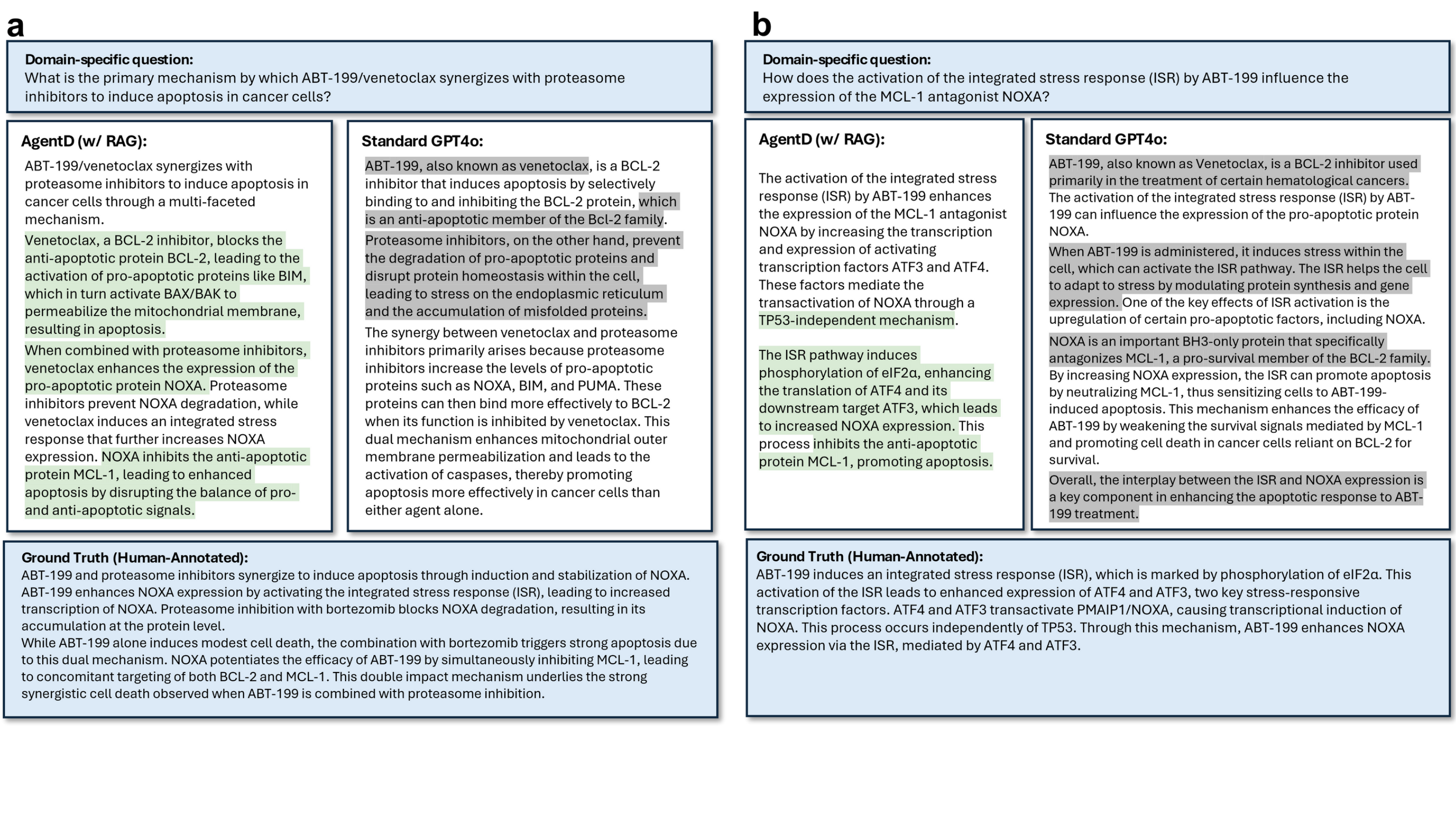} 
\caption{Comparison of question-answering performance between AgentD with RAG and the Vanilla GPT-4o model.  
Green highlights indicate grounded, context-relevant information supported by the reference paper, while grey highlights denote generic content not directly related to the question. \textbf{a} Mechanism of ABT-199/venetoclax synergy with proteasome inhibitors.  
\textbf{b} Effect of ABT-199–induced integrated stress response on NOXA expression.}
\label{fig:qna}
\end{figure*}

We evaluated this capability using a representative study by Weller et al., which describes how venetoclax activates the integrated stress response (ISR), leading to NOXA upregulation and MCL-1 inhibition~\cite{qna_ex1}. Based on this paper, we formulated domain-specific questions and compared the responses generated by AgentD with RAG to those from the vanilla GPT-4o model (Figure~\ref{fig:qna}).

AgentD consistently generates more detailed and mechanistically accurate responses. For example, when asked about the synergy mechanism between venetoclax and proteasome inhibitors, the RAG-augmented answer includes key elements such as ISR activation, ATF3/ATF4-mediated transcription of NOXA, and downstream effects involving BIM, BAX/BAK activation, and mitochondrial membrane permeabilization,closely aligning with the mechanistic explanation presented in the source paper (Figure~\ref{fig:qna}a). In contrast, vanilla GPT-4o provides a more generic explanation, omitting critical components such as the ISR pathway and transcriptional regulators.

A similar pattern is observed when inquiring about how ABT-199 influences NOXA expression through ISR. AgentD correctly references eIF2\(\alpha\) phosphorylation, TP53 independence, and the ATF3/ATF4 regulatory cascade. Although the GPT-4o response is fluent, it lacks these essential mechanistic details. As shown in Figure~\ref{fig:qna}, AgentD’s context-aware answers are highlighted in green, whereas GPT-4o’s generic responses appear in grey.

\begin{table}[hbt!]
\centering
\caption{Comparison of BERTScores between AgentD (with RAG) and vanilla language models (without RAG). Values are averaged across six question--answer pairs. Results for two representative pairs are shown in Figure~\ref{fig:qna}, and additional examples are provided in Supplementary Information Section 1.}
\begin{tabular}{lccc}
\hline
\textbf{Method} & \textbf{Precision} & \textbf{Recall} & \textbf{F1-score} \\
\hline
AgentD (GPT-4o, w/ RAG)        & 0.719 $\pm$ 0.048 & 0.686 $\pm$ 0.064 & 0.701 $\pm$ 0.046 \\
Vanilla GPT-4o (w/o RAG)       & 0.609 $\pm$ 0.018 & 0.661 $\pm$ 0.039 & 0.633 $\pm$ 0.024 \\
AgentD (DeepSeek-Chat, w/ RAG) & 0.685 $\pm$ 0.064 & 0.717 $\pm$ 0.042 & 0.700 $\pm$ 0.051 \\
Vanilla DeepSeek-Chat (w/o RAG)& 0.645 $\pm$ 0.021 & 0.698 $\pm$ 0.020 & 0.670 $\pm$ 0.012 \\
\hline
\end{tabular}
\label{tab:scibert}
\end{table}

To quantitatively assess these observed differences, we measured the semantic similarity between responses generated with and without RAG and human-annotated reference answers derived from the literature using BERTScore~\cite{Zhang2019-nn}. The reference answers were curated by the authors through careful reading of the relevant source papers~\cite{qna_ex1, Leonetti2019}. BERTScore quantifies contextual similarity by comparing the embeddings of model-generated and reference texts. We employed SciBERT~\cite{Beltagy2019-eb} as the embedding encoder because it is pre-trained on a large corpus of scientific publications from Semantic Scholar, capturing richer representations of biomedical and technical terminology than general-domain encoders such as RoBERTa~\cite{Li2025-iy}.

As shown in Table~\ref{tab:scibert}, AgentD consistently outperforms the corresponding vanilla language models across both GPT-4o and DeepSeek-Chat backends. For GPT-4o, the use of RAG increases the mean F1-score from 0.63 $\pm$ 0.01 to 0.70 $\pm$ 0.05, primarily driven by a substantial gain in Precision (0.72 vs. 0.61). This indicates that AgentD more effectively suppresses unsupported or generic statements during question answering. A similar trend is observed with DeepSeek-Chat: AgentD improves the F1-score from 0.67 $\pm$ 0.01 in the vanilla model to 0.70 $\pm$ 0.05, again with a noticeable increase in Precision (0.69 vs. 0.65). These results show that integrating RAG and structured reasoning enables AgentD to produce more faithful and literature-grounded responses.


\subsection{Seed Molecule Generation}

The initial library of molecules serves as a critical foundation for exploring chemical space and identifying candidates for further optimization. To construct this seed molecule pool, AgentD leverages two complementary generation strategies. REINVENT enables \textit{de novo} molecule generation, allowing for broad and unbiased exploration of chemical space without the need for an input structure. Mol2Mol performs conditional generation, producing analogs that are structurally similar to a specified input molecule. In our workflow, the existing drug identified during the data extraction phase, such as venetoclax, is used as input for Mol2Mol, enabling the agent to focus on chemically relevant and biologically meaningful regions of the search space.

\begin{figure*}[!htb] 
\centering
\includegraphics[width=0.82\textwidth]{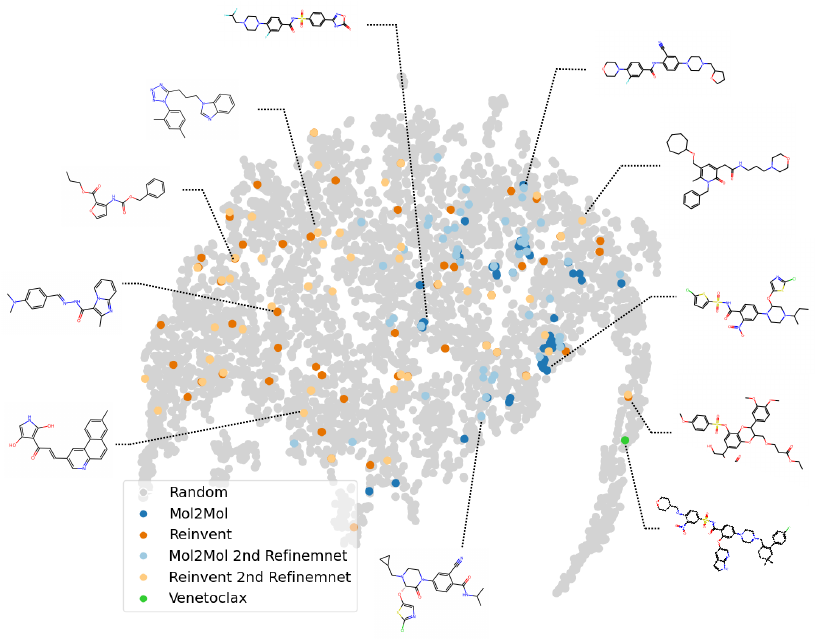} 
\caption{t-SNE visualization of chemical space, based on Mordred-derived molecular descriptors~\cite{mordred}, including seed molecules generated by AgentD (Mol2Mol, REINVENT), refined molecules after property-aware optimization, 5,000 randomly sampled compounds from ChEMBL, and the reference drug venetoclax.}
\label{fig:tsne}
\end{figure*}

After retrieving the SMILES of the known drug, AgentD automatically integrates it into a configuration file and executes both REINVENT and Mol2Mol to generate the initial seed molecules. As shown in Figure~\ref{fig:tsne}, REINVENT-generated molecules are widely distributed across the chemical space, reflecting its exploratory capabilities, whereas Mol2Mol-generated molecules are more tightly clustered around the input molecule, enabling targeted exploration near known active scaffolds. This seed library serves as the starting point for downstream tasks such as property prediction, refinement, and structure generation.

\subsection{Property-Aware Molecular Refinement}

To enhance the quality of the seed molecule pool, AgentD performs property-driven refinement of SMILES structures by identifying and addressing limitations that compromise their drug-likeness. Using the Deep-PK model API~\cite{deep_pk}, the agent predicts 67 ADMET properties, as well as 7 general physicochemical properties. A complete list of predicted ADMET properties is provided in Supplementary Table S2. In parallel, AgentD uses the BAPULM model~\cite{bapulm} to estimate binding affinity (pKd).

\begin{figure*}[!htbp]
\centering
\includegraphics[width=0.99\textwidth]{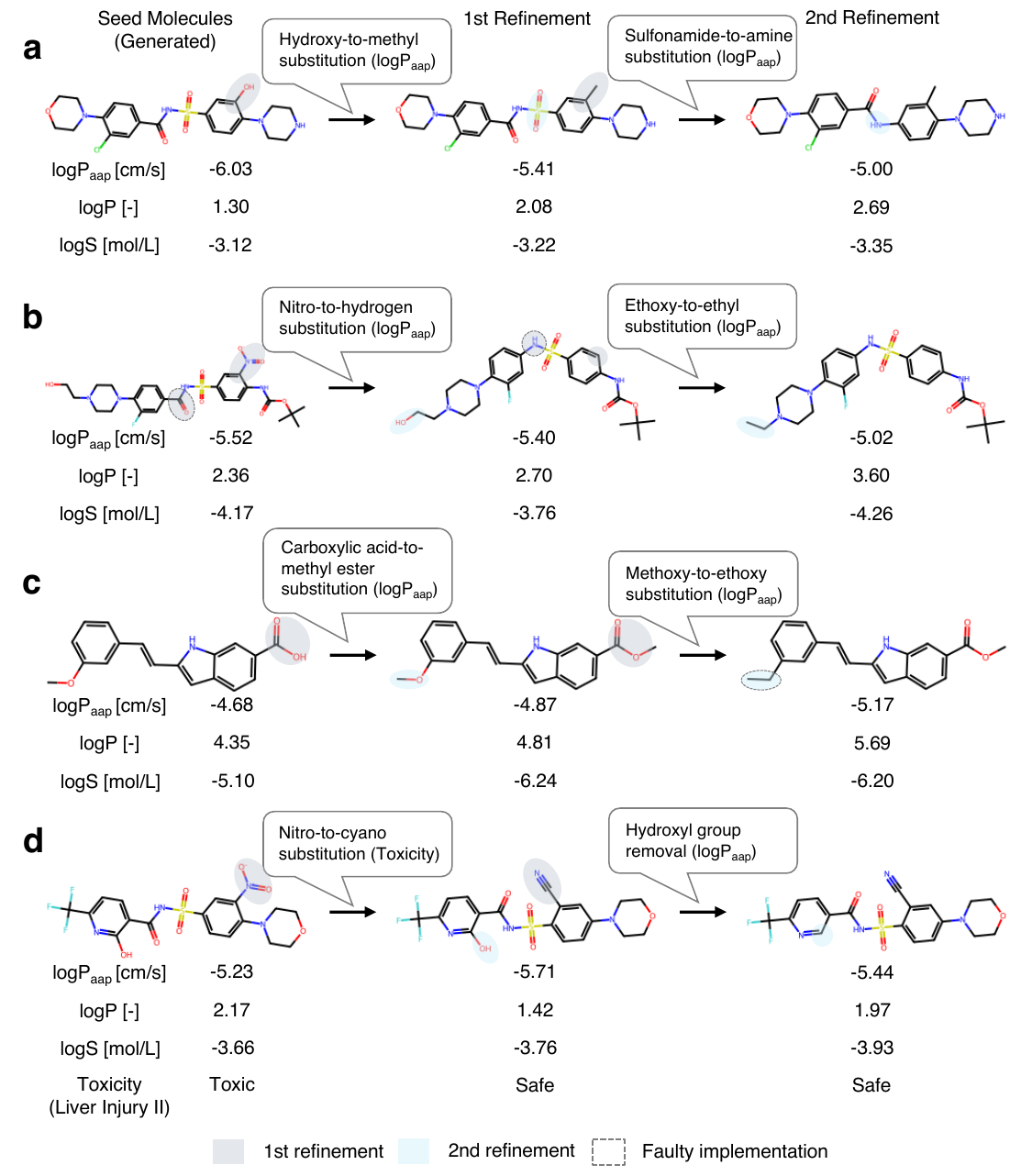} 
\caption{Property-aware molecular refinements by AgentD.  
\textbf{a} Iterative changes improve permeability.  
\textbf{b} An unintended modification leads to improved permeability.  
\textbf{c} An intended modification fails to improve permeability.  
\textbf{d} Iterative changes improve both toxicity and permeability.}
\label{fig:refine_example}
\end{figure*}

Based on these predictions, AgentD identifies unfavorable molecular properties relevant to drug development, such as low permeability or high toxicity, and proposes targeted structural edits to improve them. A comprehensive list of weakness-associated properties is provided in Supplementary Table S3. We demonstrate this refinement process over two iterations, beginning with an initial set of 100 molecules. These iterations produced 99 and 95 additional valid SMILES, respectively, which were added to the candidate pool. For instance, a molecule is initially flagged for poor permeability due to a low predicted $\log P_{\text{app}}$ value (Figure~\ref{fig:refine_example}a). In response, AgentD suggested replacing a hydroxyl group with a methyl group in the first round, followed by substituting a sulfonamide with an amine in the second - both changes contributed to improved predicted permeability. Figure~\ref{fig:refine_example}d illustrates another case of successful refinement, where the targeted weak properties, such as toxicity and permeability, were improved at each step.

However, refinements are not always successful or precise. Two common failure modes are observed: (i) unintended SMILES modifications that diverge from the agent’s stated intent, and (ii) correctly executed modifications that fail to improve the target property. For case (i), in Figure~\ref{fig:refine_example}b, the agent intended a nitro-to-hydrogen substitution and successfully applied it, but also unintentionally removed a carbonyl group; nonetheless, the predicted permeability improved. Additionally, in Figure~\ref{fig:refine_example}c, an intended methoxy-to-ethoxy substitution was improperly implemented, resulting in the loss of an oxygen atom and decreased permeability. For case (ii), even when the modification is applied correctly, the desired outcome may not be achieved, as seen in the first-round refinement of Figure~\ref{fig:refine_example}c, where the permeability worsened despite the intended structural change.


\begin{figure*}[!htb] 
\centering
\includegraphics[width=0.99\textwidth]{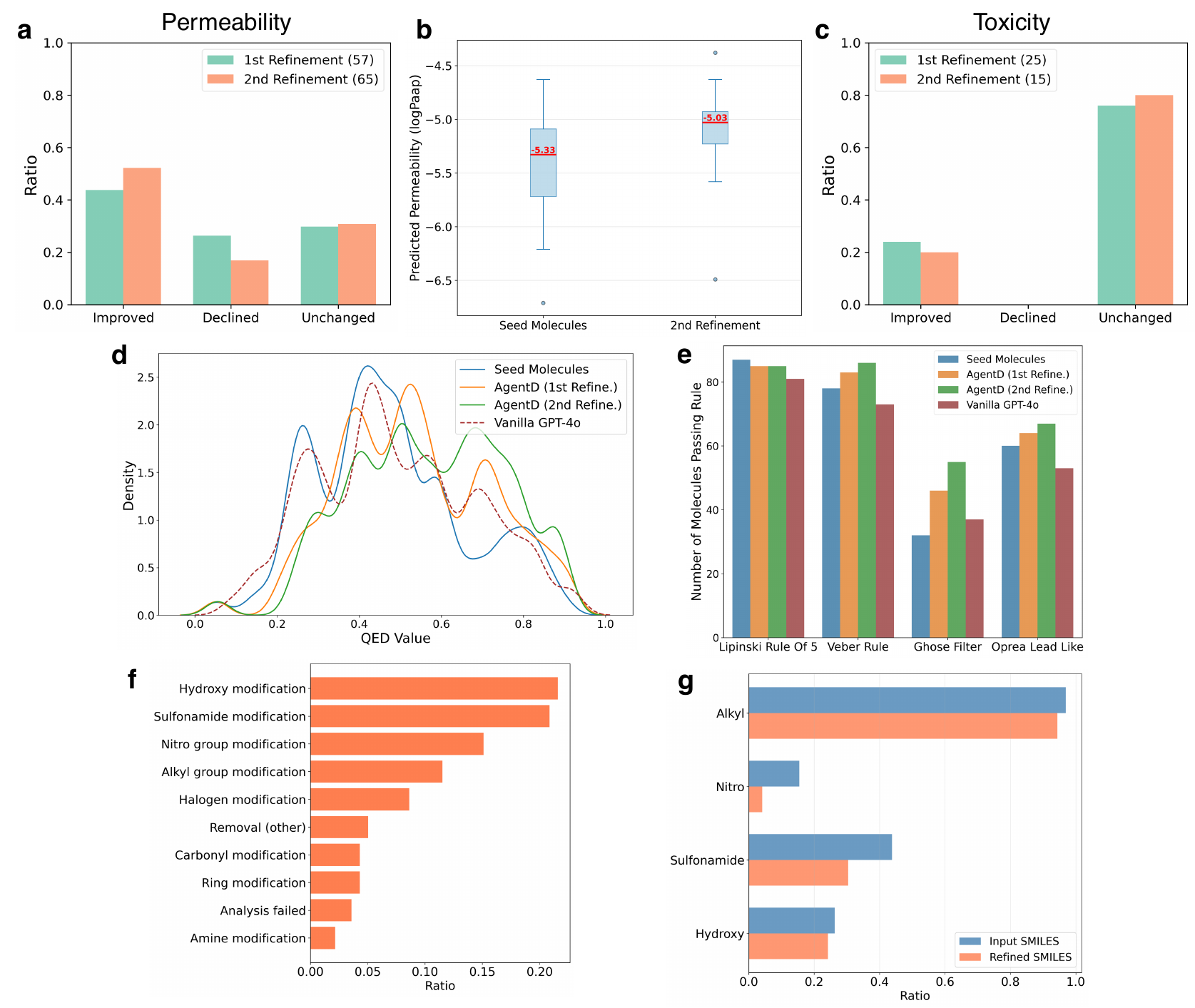} 
\caption{Property-aware molecular refinement results.  
\textbf{a} Proportion of molecules with improved, declined, or unchanged permeability (logP\textsubscript{app}). 
\textbf{b} Boxplot of logP\textsubscript{app} for entries showing improvement. 
\textbf{c} Molecular refinement patterns for molecules with improved permeability. 
\textbf{d} QED distribution shifts across iterations (vanilla GPT-4o indicates refinement solely through direct queries to GPT-4o). 
\textbf{e} Counts of molecules satisfying empirical drug-likeness rules (definitions in Methods).
\textbf{f} Fraction of molecules modified for permeability; ``Analysis failed" indicates unresolved substructure differences by the substructure algorithm.  
\textbf{g} Functional-group frequencies in input vs. refined molecules across iterations (input = input for refinement; refined = post-refinement outputs).}
\label{fig:refinement}
\end{figure*}

In the first refinement round, 57 out of the initial 100 molecules were identified as having low permeability. Among these, approximately 44\% showed improved $\log P_{\text{app}}$ values after refinement, 26\% declined, and 30\% remained unchanged (Figure~\ref{fig:refinement}a). In the second round, 52\% improved, 17\% declined, and 31\% were unchanged. For the molecules that showed target property improvement, the median $\log P_{\text{app}}$ increased from –5.33 to –5.03 (Figure~\ref{fig:refinement}b). Although the magnitude of change is modest, the upward trend is clearly visible within just two iterations. Toxicity improvements were relatively more limited: 24\% and 20\% of high-toxicity molecules showed improvement in the first and second rounds, respectively, with most remaining toxic despite modification (Figure~\ref{fig:refinement}c).

Despite occasional local failures, as illustrated in Figure~\ref{fig:refine_example}b and c, the overall impact of refinement is positive across the molecule pool. This is supported by increased QED (Quantitative Estimate of Drug-likeness) scores, which reflect overall drug-likeness based on empirical physicochemical properties \cite{qed}. As shown in Figure~\ref{fig:refinement}d, the distribution of high-QED molecules expanded with each refinement iteration: the number of molecules with QED $>$ 0.6 increased from 34 in the original set to 49 after the first update, and to 55 after the second.

As shown in Figure~\ref{fig:refinement}e, LLM-based refinement increased the proportion of molecules meeting empirical drug-likeness and lead-likeness criteria, including Lipinski’s Rule of Five, Veber’s rule, Ghose’s filter, and Oprea’s lead-likeness filter~\cite{lipinski2001experimental, veber2002molecular, ghose1999knowledge, oprea2001lead}. These rules assess different aspects of molecular suitability: Lipinski’s and Veber’s rules emphasize oral bioavailability and permeability, Ghose’s filter captures physicochemical ranges observed in marketed drugs, and Oprea’s filter identifies compounds suitable as lead-like starting points for optimization. After refinement, more molecules satisfied Veber’s rule (78 to 86) and Oprea’s filter (60 to 67). The most notable improvement was observed for Ghose’s filter, with the number of passing molecules increasing from 32 to 55 ($\sim$72\% increase). These results indicate that refinement helped expand the chemical space toward drug-like and lead-like regions, particularly in terms of Ghose’s statistical descriptors. Compliance with Lipinski’s rule remained largely unchanged; this outcome is further discussed in the Discussion section.

AgentD tends to prioritize edits to polar or electron-withdrawing groups, such as hydroxyls, sulfonamides, and nitro groups, when attempting to improve permeability-related properties. Figure~\ref{fig:refinement}f shows the distribution of SMILES modifications for entries with permeability as the target weakness property. The most frequent edits involved hydroxyl groups (30 cases, 15.5\%), followed by sulfonamide substitutions (29 cases, 15.0\%) and nitro group changes (21 cases, 10.8\%). Other commonly modified motifs included alkyl groups (8.3\%) and halogens (6.2\%), and together these five categories account for 55.7\% of all refinements. Also, AgentD’s refinement behavior does not merely reflect functional-group prevalence. For instance, despite nitro groups being less common than alkyl, sulfonamide, or hydroxyl groups, nitro edits still represent a substantial portion of modifications (10.8\%; Figure~\ref{fig:refinement}g).

\subsection{3D Structure Generation}

After refining the seed molecule pool through property-aware modifications, we applied empirical filtering criteria to select candidates for 3D structural evaluation. The purpose of this filtering scheme is not to identify a final drug candidate, but to illustrate the agent’s capability to perform key tasks in the drug discovery pipeline. In our workflow, molecules are shortlisted if they (i) satisfy Oprea’s lead-likeness filter, (ii) pass at least two of the three drug-likeness rules (Lipinski’s Rule of Five, Veber’s rule, and Ghose’s filter), (iii) achieve a QED score above 0.55, and (iv) have a predicted pKd value greater than 6.0. 

\begin{figure*}[!htb] 
\centering
\includegraphics[width=0.99\textwidth]{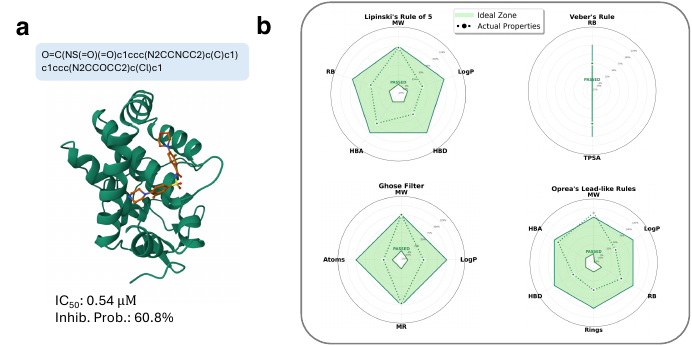} 
\caption{Protein–ligand structure generation and evaluation. \textbf{a} A representative 3D complex predicted by Boltz-2, with estimated IC\textsubscript{50} and inhibitor probability. \textbf{b} Radar plot showing empirical rule-based evaluation. Green dotted lines indicate the thresholds for each rule, while the purple solid line represents the properties of the selected molecule. See Methods and Table~S4 for criteria.}
\label{fig:structure1}
\end{figure*}

The molecule shown in Figure~\ref{fig:structure1} satisfies the lead-likeness requirement, passes all three drug-likeness rules, has a QED score of 0.68, and achieves a predicted pKd of 6.18. This is simply a representative candidate for further analysis. It is important to note that these empirical rules may not always align, and molecules rarely satisfy all of them simultaneously. In fact, only one molecule in our set passed all three rules, which we selected as an illustrative example for 3D visualization. Additional examples are provided in Supplementary Figure S7.

AgentD uses Boltz-2 to generate 3D protein–ligand complex structures using the ligand’s SMILES and the target protein’s FASTA sequence. These structures provide a foundation for more rigorous downstream evaluations, such as molecular docking, MD simulations, and free energy perturbation analyses, to assess binding stability, conformational flexibility, and interaction specificity under biologically relevant conditions. In addition to 3D structure, Boltz-2 returns estimated binding metrics, including logIC\textsubscript{50} values and inhibitor probabilities. In the example shown, the predicted IC\textsubscript{50} (0.54 µM) and inhibitor probability indicate good sub-micromolar binding affinity, making the molecule a suitable candidate for further validation using molecular dynamics simulations.

\subsection{Comparison Across Language Models}

AgentD can operate with a variety of LLMs, including those from OpenAI, Anthropic, and DeepSeek. In practice, the framework is optimized for GPT-4o. We tested AgentD using GPT-4o, Claude-3.7-Sonnet, and DeepSeek-Chat \cite{openai2023gpt4, anthropic2024claude, deepseekai2025, lu2024deepseekvl}. DeepSeek-Chat, an open-source model, successfully executed all six tasks and demonstrated behavior broadly comparable to GPT-4o. In contrast, Claude-3.7-Sonnet exhibited failure modes in two tasks: domain-specific question answering and molecular refinement. 

In the question-answering task, Claude-3.7-Sonnet encountered syntax errors when parsing PDF files. AgentD, powered by GPT-4o, achieved higher BERTScores than AgentD with DeepSeek-Chat (Table~\ref{tab:scibert}). Among the vanilla language models, DeepSeek-Chat outperformed GPT-4o, highlighting that the performance gain provided by AgentD is more pronounced when using GPT-4o.

\begin{figure*}[!htb] 
\centering
\includegraphics[width=0.99\textwidth]{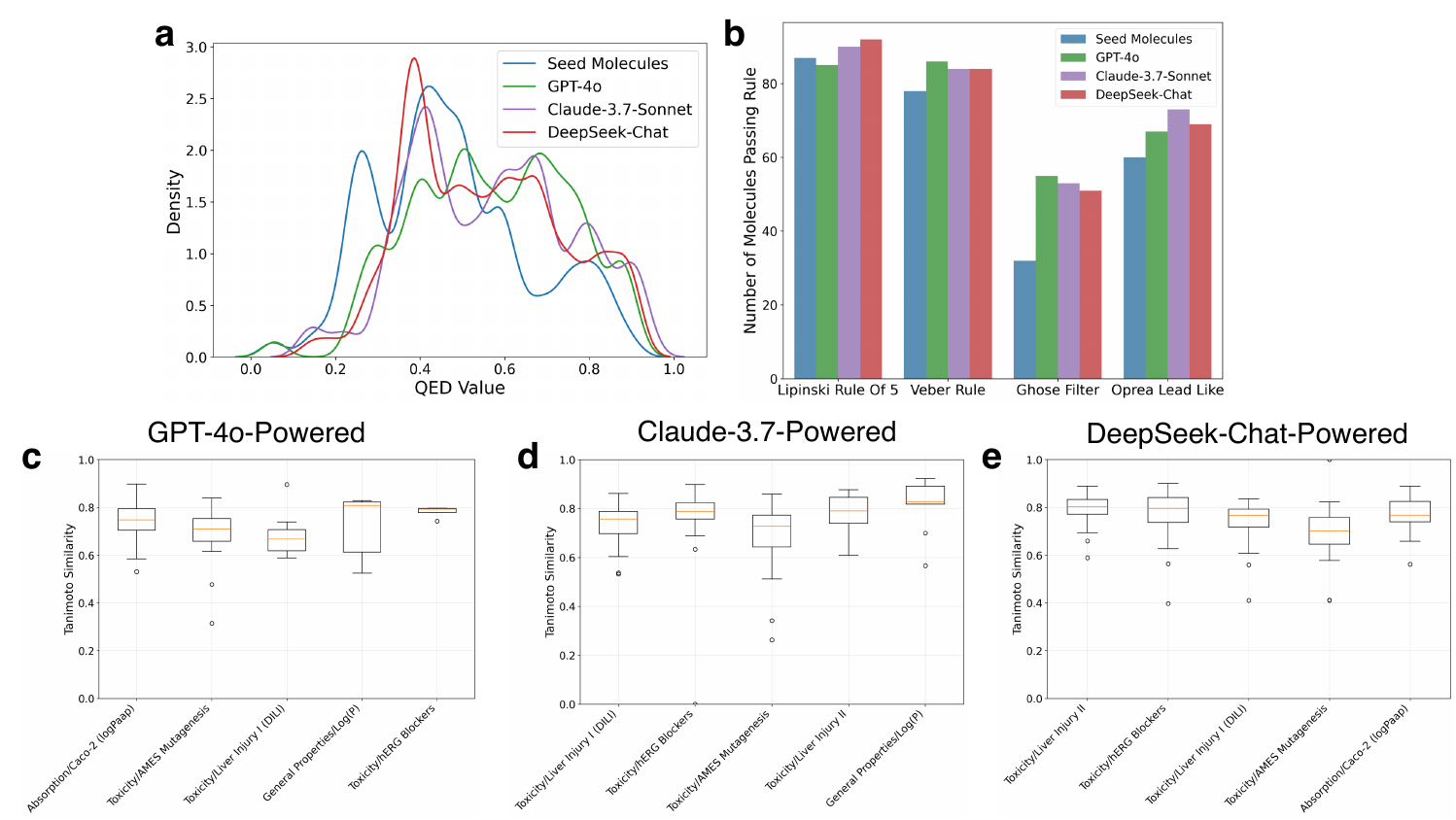} 
\caption{Comparison of AgentD performance across different language models. 
\textbf{a} Distribution of QED values before and after two iterations of property-aware molecular refinement for AgentD powered by GPT-4o, Claude-3.7-Sonnet, and DeepSeek-Chat. 
\textbf{b} Histogram showing the number of molecules passing empirical lead- and drug-like rules. 
\textbf{c--e} Tanimoto similarity between the original and updated SMILES for the top five most frequently targeted weakness properties for AgentD powered by GPT-4o (\textbf{c}), Claude-3.7-Sonnet (\textbf{d}), and DeepSeek-Chat (\textbf{e}).}
\label{fig:compare_lms}
\end{figure*}

In the molecular refinement task, AgentD with Claude-3.7-Sonnet occasionally became trapped in repeated attempts to generate valid updated SMILES, despite explicit instructions to avoid infinite retries (see Supplementary Figure S4). For the quantitative comparison of property-aware molecular refinement shown in Figure~\ref{fig:compare_lms}, we report results from a verified successful run of Claude-3.7-Sonnet to ensure a fair comparison.

All three model cases improved the QED of seed molecules after two iterations of property-aware refinement (Figure~\ref{fig:compare_lms}a). Starting from an average QED of 0.47, AgentD, powered by GPT-4o, increased the mean value to 0.57, Claude-3.7-Sonnet increased it to 0.59, and DeepSeek-Chat increased it to 0.55. While all models demonstrated meaningful improvements, DeepSeek-Chat achieved a slightly smaller gain relative to the others. The seed set contained 23.0\% of molecules with QED \textgreater 0.6, which increased to 45.7\% with GPT-4o, 44.9\% with Claude-3.7-Sonnet, and 40.0\% with DeepSeek-Chat after the second refinement step.

The rule-based evaluation shows a similar trend across the three LLMs, with the most evident increase observed in the number of molecules passing the Ghose filter (Figure~\ref{fig:compare_lms}b). Claude-3.7-Sonnet and DeepSeek-Chat do not exhibit the slight deterioration in Lipinski’s Rule of Five observed with GPT-4o, although both models yield slightly fewer molecules passing Veber’s rule.

Each language model exhibits some variability in the selection of ``weakness" properties during molecular refinement. Figures~\ref{fig:compare_lms}c–e report the Tanimoto similarity~\cite{tanimoto, tanimoto2} between the original SMILES and the updated SMILES for the top five most frequently targeted weakness properties. For AgentD powered by GPT-4o, the top five properties were logP\textsubscript{app}, AMES mutagenicity, liver injury I toxicity, logP, and hERG blocker toxicity, with median similarities of 0.71, 0.71, 0.67, 0.81, and 0.79, respectively. Despite model-specific differences, the selected properties largely arise from a consistent subset of the 75 property endpoints, indicating broadly aligned refinement priorities. AgentD with Claude-3.7-Sonnet prioritized liver injury I toxicity, hERG blocker toxicity, AMES mutagenicity, liver injury II toxicity, and logP, with median similarities of 0.76, 0.79, 0.73, 0.79, and 0.83. For AgentD with DeepSeek-Chat, the top five properties were liver injury II toxicity, hERG blocker toxicity, liver injury I toxicity, AMES mutagenicity, and logP\textsubscript{app}, with median similarities of 0.80, 0.80, 0.77, 0.70, and 0.77. Although all three models generally target similar toxicity- and permeability-related weaknesses, the specific ranking of properties and the degree of structural change.

\section{Discussion}

\subsection{Property Awareness Improves Drug-Likeness}

Weakness-property awareness substantially improves drug-likeness, even though drug-likeness is not explicitly used as an optimization target. Refinement via a vanilla LLM, without property-aware guidance, yields little measurable improvement (Figure~\ref{fig:refinement}d–e). In contrast, incorporating weakness-property information leads to pronounced gains, indicating that providing explicit context about what is deficient in a molecule enables more targeted and pharmacologically meaningful edits. This functions similarly to chain-of-thought reasoning: giving the LLM stepwise situational cues (e.g., ``this is the weakness, so modify accordingly") helps steer its structural decisions in a more directed and interpretable manner.

AgentD’s refinement pattern also does not simply mirror the prevalence of functional groups in the dataset (Figure~\ref{fig:refinement}f–g). Instead, the model displays distinct preferences in the types of chemical transformations it applies, suggesting that the observed patterns arise from its learned reasoning rather than from dataset-driven frequencies.

The benefit of property-aware refinement is consistent across all LLMs used as the agent’s core engine (Figure~\ref{fig:compare_lms}a–b). Whether using GPT-4o, Claude-3.7-Sonnet, or DeepSeek-Chat, the refinement strategy reliably increases drug-likeness. This indicates that the performance gains stem from the refinement paradigm itself rather than from model-specific idiosyncrasies.

\subsection{Challenges in Molecular Refinement}

The property-aware molecular refinement module also shows limitations. Molecular optimization remains intrinsically complex due to interdependencies among properties. For example, in Figure~\ref{fig:refine_example}a and b, predicted apparent permeability (logP\textsubscript{app}) is influenced by lipophilicity (logP), consistent with Overton’s rule~\cite{overton}, whereas aqueous solubility (logS) does not follow a consistent trend and may either increase or decrease. These observations highlight the complex, non-linear interplay among lipophilicity, solubility, and permeability. Similarly, the first-round refinement in Figure~\ref{fig:refine_example}d mitigates predicted liver toxicity but worsens permeability. These trade-offs highlight the fundamental challenge of multi-property optimization in drug design. These findings highlight the importance of incorporating multi-objective optimization in future iterations to better balance competing pharmacological goals. Despite these complexities, AgentD achieves a net improvement in overall drug-likeness across the molecule pool, as shown in Figure~\ref{fig:refinement}c and d. 

Molecular optimization is also target-dependent, as multiple interacting physicochemical properties influence outcomes. Across the three therapeutic targets, BCL-2, JAK-2, and thrombin, drug-likeness improves overall, as reflected by QED distributions and rule-based filters, though the improvement patterns differ. For BCL-2 and thrombin, the largest gains occur under the Ghose filter, while Lipinski’s Rule of Five shows only minor improvement, likely because oral bioavailability was rarely selected as a weakness (Supplementary Table S3). By contrast, JAK-2 shows stronger gains under Lipinski’s criteria but minimal change under the Ghose filter. These differences highlight the intrinsic complexity of molecular refinement, where multiple interdependent properties are convoluted.

\subsection{Effective Retrieval and Generation}

The agent reliably retrieves protein and ligand data from both structured databases and web sources. While our current implementation focuses on targeted extraction, future extensions could incorporate large-scale literature mining and relevance scoring to support broader knowledge synthesis. 

For domain-specific question answering, AgentD’s RAG-augmented responses consistently outperform standard LLM outputs in contextual accuracy and mechanistic completeness (Table~\ref{tab:scibert}). Because the effectiveness of RAG depends strongly on the quality of retrieved literature, expanding retrieval beyond open-access sources may further enhance performance. Our current implementation uses a deliberately minimal retrieval pipeline, without filters on publication date, venue quality, or citation metrics, yet it still provides clear gains in factual grounding and mechanistic fidelity. These findings demonstrate that even a raw, loosely constrained RAG setup can significantly improve scientific question answering. 


For molecule generation, AgentD integrates external models such as REINVENT and Mol2Mol using natural language-driven configuration updates, enabling seamless integration of new tools with minimal customization. By leveraging a language model to interpret and modify configuration files based on high-level instructions, the system can flexibly switch between different generative models or sampling strategies without manual code changes. This approach removes the need for model-specific hardcoded logic and allows for rapid adaptation to evolving architectures and file structures. The modular design of AgentD further ensures its long-term adaptability as generative modeling techniques continue to advance.

The structure generation task complements the pooling and refinement process by enabling 3D protein-ligand complex generation, which provides critical input for structure-based studies. Boltz-2 captures meaningful trends in ligand binding behavior, but its IC\textsubscript{50} and inhibitor probability predictions should be interpreted as rough approximations that are useful for initial prioritization rather than as substitutes for detailed structure-based evaluation.

\subsection{Limitations and Future Directions}

Despite its modular design and encouraging end-to-end performance, AgentD has several limitations. The workflow relies on external predictive tools such as Deep-PK, BAPULM, and Boltz-2, each of which carries its own error profile and uncertainties. Inaccuracies in these upstream modules can propagate downstream, affecting both refinement decisions and compound prioritization. While a fault-tolerance scheme has been implemented for Deep-PK, similar safeguards are not yet in place for other components.

Another source of variability arises from the inherently non-deterministic nature of weakness property identification during refinement. Because the agent may select different properties as optimization targets across runs, exact reproduction of a single refinement is challenging. For this reason, we analyze outcomes at the group level rather than on a per-run or per-molecule basis. Additionally, the RAG module is currently configured in a minimal form, performing unfiltered retrieval without consideration of venue quality or citation strength.

Future iterations of AgentD could address these limitations by integrating uncertainty-aware property prediction, ensemble scoring, and confidence-weighted refinement to reduce sensitivity to individual tool errors. More sophisticated retrieval strategies, such as relevance ranking or knowledge graph integration, would strengthen literature grounding, while feedback from molecular dynamics or free-energy calculations could provide physically validated guidance for structural optimization.

\section{Conclusion}

In this study, we introduced AgentD, a modular, LLM-powered agent framework for automating and streamlining key tasks of the drug discovery pipeline. By integrating language model reasoning with domain-specific tools and databases, AgentD can: (i) retrieve relevant protein and compound data from structured databases and web sources; (ii) answer domain-specific scientific questions grounded in literature; (iii) generate diverse, context-aware molecules using both de novo and conditional models; (iv) predict pharmacologically relevant properties; (v) refine molecular representations through iterative, property-aware optimization; and (vi) construct protein–ligand complex structures for downstream simulations. 

AgentD marks a step toward general-purpose, AI-driven scientific agents for therapeutic discovery. Its modular architecture supports seamless integration of new models and tools, ensuring continued adaptability as technologies evolve. Future extensions may enhance its capabilities through autonomous molecular dynamics simulations for structural validation, multi-objective molecular optimization, and the generation of candidates conditioned on specific pharmacological profiles.

\section{Methods}
\subsection{Large Language Model Agent}
Our framework primarily employs OpenAI's GPT-4o model~\cite{openai2023gpt4} as the main language model for all AgentD modules. GPT-4o is an optimized version of GPT-4, part of the Generative Pretrained Transformer family~\cite{radford2018improving}. In addition, we integrated Anthropic's Claude-3-7-sonnet-20250219 and DeepSeek-Chat models. All models are accessed via their respective APIs and loaded dynamically using the LangChain library~\cite{chase2022langchain}, which simplifies prompt management, response processing, and integration with external tools. For all API calls, no local GPU resources are required; however, for tool-specific tasks such as binding affinity prediction, SMILES generation, and protein–ligand structure generation, computations are performed on a single NVIDIA RTX 2080 Ti GPU. All models are used with default settings provided by the API.

\subsection{Database}
We utilize two established bioinformatics resources, UniProt and ChEMBL, to access protein sequence data and small molecule representations, respectively. UniProt provides high-quality annotated protein sequences between species, enabling access to functionally characterized targets of therapeutic relevance \cite{uniprot}. For each therapeutically relevant protein target (e.g., \textit{EGFR} or \textit{TP53}), the agent queries UniProt’s REST API, restricted to the \textit{Homo sapiens} taxonomy (organism ID: 9606) to retrieve the corresponding amino acid sequence in FASTA format. When existing drugs are available for the target, we subsequently query ChEMBL to obtain SMILES representations of these known compounds\cite{chembl}. This ChEMBL query step is only executed if existing drugs for the target are identified through prior web searches; otherwise, this step is skipped.

\subsection{Retrieval-Augmented Generation}

AgentD incorporates a lightweight Retrieval-Augmented Generation (RAG) workflow to provide literature-grounded responses during domain-specific question answering. In its current implementation, literature retrieval is intentionally kept minimalistic: the agent generates search keywords, and papers are retrieved solely through the Serper API using these keywords, without applying explicit filters on publication year, venue quality, or citation count. As a result, the retrieved documents may vary in quality and relevance. 

Once retrieved, PDFs are processed with a \texttt{CharacterTextSplitter} (chunk size 1000 characters, 50-character overlap) to maintain contextual continuity across segments. Each chunk is embedded using \texttt{EMBEDDING\_MODEL} and stored in a FAISS vector index to allow uniform and efficient retrieval. During the question-answering phase, the LLM performs similarity search over the FAISS index to identify the most relevant text segments, which are then supplied as grounding context for response generation.

\subsection{ADMET Properties}
ADMET properties encompass Absorption, Distribution, Metabolism, Excretion, and Toxicity characteristics that are critical for drug development~\cite{admet1, admet2}. These pharmacologically relevant descriptors determine how a compound is absorbed into the bloodstream, distributed across tissues, metabolized by enzymatic systems, eliminated from the body, and whether it poses potential toxic effects. Early prediction of ADMET properties is essential for prioritizing compounds with favorable biopharmaceutical and safety profiles, thereby reducing downstream attrition during drug development~\cite{vanDeWaterbeemd2001,leeson2012druglike}.

As part of this multi-stage assessment, we integrated Deep-PK, an ADMET prediction framework that operates on SMILES input and internally employs a Message Passing Neural Network (MPNN) to capture the atomic and topological features of each molecule~\cite{deep_pk}. Candidate ligands, whether recovered from ChEMBL or generated, are submitted to the Deep-PK REST API via their SMILES strings, and the resulting ADMET profiles are parsed to prioritize compounds with favorable pharmacokinetic and safety attributes. The complete set of predicted properties is summarized in Supplementary Table S2.

A tolerance mechanism was implemented for Deep-PK API requests. After submitting SMILES, the system polls for job completion at fixed intervals (30~s) up to a maximum wait time (500~s). API errors or missing job IDs are logged and returned without interrupting the pipeline, and running jobs are retried until completion or timeout.

\subsection{Binding Affinity}
Binding affinity to the biological target is a fundamental determinant of therapeutic potential, serving as a key predictor of drug potency and selectivity. We employ two complementary strategies for affinity prediction. In the property prediction task, the sequence-based BAPULM model employs a dual encoder architecture to estimate the dissociation constant ($K_d$) from protein amino acid sequences and ligand SMILES representations. It integrates two domain-specific pre-trained language models: ProtT5-XL-U50 for proteins and MolFormer for small molecules~\cite{prot5, molformer}. Each encoder generates latent embeddings tailored to its respective input, which are then projected into a shared latent space using learnable feedforward projection heads. Subsequently, a predictive head processes these joint representations to estimate the binding affinity, reported as $pK_d = -\log_{10}(K_d)$\cite{bapulm}. Given its computational efficiency and reliance solely on sequence-level inputs, it is well-suited for early-stage screening of large ligand libraries based on predicted binding affinity.

In the structure generation task, the structure-based model Boltz-2 begins with the same sequence and SMILES input but internally generates 3D protein–ligand complex structures. These conformations are then used to predict the half-maximum inhibitory concentration (IC$_{50}$), reflecting the inhibitory efficacy of a compound in biochemical assays~\cite{wang2023computing}. In addition to regression-based affinity values, Boltz also outputs inhibitor probability scores, indicating the likelihood that a given ligand acts as an active binder. Although $K_d$ and IC$_{50}$ originate from different experimental setups, they are related through the Cheng-Prusoff equation \cite{cheng1973relationship}. 

\begin{equation}
K_i = \frac{\mathrm{IC}_{50}}{1 + \frac{[S]}{K_m}}
\label{eq:cheng_prusoff}
\end{equation}

where $K_i$ approximates $K_d$ under certain biochemical conditions.

\subsection{SMILES Generation}

We utilize two molecular generators: REINVENT for de novo molecular design and Mol2Mol for molecular optimization~\cite{reinvent4, reinvent2, mol2mol}. Both generators utilize recurrent neural networks and transformer architectures and are embedded within machine learning optimization algorithms, including reinforcement learning and transfer learning~\cite{reinvent4, rl_drug}. Both models were implemented using the REINVENT4 package and configured to generate one molecule per input SMILES (`num\_smiles = 157'), while retaining only unique molecules (`unique\_molecules = true') with canonicalized SMILES to eliminate duplicates.

REINVENT performs \textit{de novo} molecular generation using sequence-based models that capture SMILES token probabilities in an autoregressive manner. The models are trained via teacher-forcing on large SMILES datasets to learn chemical syntax and generate valid molecules~\cite{Williams1989}. 

Mol2Mol performs conditional generation, accepting input SMILES strings and generating structurally similar molecules within user-defined similarity constraints~\cite{mol2mol}. The transformer-based model was trained on over 200 billion molecular pairs from PubChem with Tanimoto similarity $\geq$ 0.50. Training employed ranking loss to directly link negative log-likelihood to molecular similarity. The model supports multinomial sampling with temperature control (set to 1.0 in this study).

\subsection{Protein-Ligand Complex Structure}

Boltz-2 is a structure-based deep learning model designed to jointly predict 3D protein–ligand complex structures by integrating protein folding and ligand binding into a unified framework~\cite{passaro2025boltz2}. The model takes as input a protein FASTA sequence and a ligand SMILES string, and simultaneously infers the full atomic conformation of the protein as well as the bound pose of the ligand within the predicted binding pocket. Unlike traditional docking pipelines that require experimentally resolved protein structures, Boltz-2 performs \textit{ab initio} structure prediction, enabling end-to-end modeling from sequence alone. 

Its architecture builds upon the Evoformer\cite{alphafold2} stack and SE(3)-equivariant transformer modules, incorporating interleaved attention mechanisms to capture long-range dependencies both within the protein sequence and between the protein and ligand~\cite{attention, wohlwend2024boltz1, passaro2025boltz2}. This allows Boltz-2 to reason over flexible ligand conformations and protein-side chain rearrangements in a physics-aware manner. By directly predicting all-atom 3D coordinates, the model enables rapid generation of realistic protein–ligand complexes suitable for downstream scoring.


\subsection{Evaluation Metrics}

To evaluate the drug-likeness, lead-likeness, and pharmacokinetic relevance of the generated molecules, we applied four widely used rule-based filters: Lipinski’s Rule of Five~\cite{lipinski2001experimental}, Veber’s Rule~\cite{veber2002molecular}, the Ghose filter~\cite{ghose1999knowledge}, and Oprea’s Lead-like Rule~\cite{oprea2001lead}. The first three rules assess general drug-likeness, while Oprea’s rule specifically addresses lead-likeness as an early-stage starting point for optimization. The detailed numerical thresholds for each rule are summarized in Table~S4 of the Supporting Information. All descriptors were computed using RDKit.

Lipinski’s Rule of Five includes five criteria: molecular weight (MW) $\leq$ 500 Da, LogP $\leq$ 5, hydrogen bond donors (HBD) $\leq$ 5, hydrogen bond acceptors (HBA) $\leq$ 10, and rotatable bonds $\leq$ 10. A molecule was considered compliant if it satisfied at least four of the five conditions. Although the original formulation included only the first four parameters, the rotatable bond threshold has since been adopted to better capture oral absorption potential.

Veber’s rule assesses polarity and flexibility using two thresholds: topological polar surface area (TPSA) $\leq$ 140 \AA$^2$ and rotatable bonds $\leq$ 10. Both conditions must be met for compliance.

The Ghose filter identifies drug-like chemical space based on statistical distributions observed in marketed drugs. Its thresholds include MW between 160–480 Da, LogP between –0.4 and 5.6, molar refractivity (MR) between 40–130, and heavy atom count between 20–70. Molecules meeting these ranges are considered to occupy favorable physicochemical space for drug development.

Oprea’s Lead-like Rule defines chemical properties suitable for ``lead" compounds—molecules that may not yet meet full drug-likeness but serve as starting points for further optimization. Criteria include MW between 200–450 Da, LogP between –1 and 4.5, HBD $\leq$ 5, HBA $\leq$ 8, rotatable bonds $\leq$ 8, and aromatic rings $\leq$ 4, with allowance for one violation. Compared to drug-like rules, these thresholds are narrower, reflecting the preference for smaller, less lipophilic molecules that can be synthetically elaborated into drug candidates.

\section*{Technology Use Disclosure}
We used ChatGPT and Claude to help with grammar and typographical corrections during the preparation of this preprint manuscript. The authors have carefully reviewed, verified, and approved all content to ensure accuracy and integrity.

\section*{Code Availability Statement}
The code that supports this study's findings can be found in the following publicly available GitHub repository: \url{https://github.com/hoon-ock/AgentD}.

\begin{acknowledgement}
The authors gratefully acknowledge support from the H. Robert Sharbaugh Presidential Fellowship.
\end{acknowledgement}

\section*{Author declarations}
\subsection*{Conflict of Interest}
The authors have no conflicts to disclose.

\subsection*{Author Contributions}
J.O., A.B.F. conceived the study, and J.O., R.S.M., A.C., N.S.A., and S.B. conducted the experiments, analyzed the data, and wrote the original draft. R.S.M., N.S.A., S.B., and A.C. contributed to experimental design and data analysis. A.B.F. supervised the project, provided funding acquisition, and reviewed the manuscript.

\bibliography{reference}

\end{document}


\tableofcontents





\section{Domain-specific Question Answering}

\begin{figure}[!htpb]
\centering
\includegraphics[width=0.75\textwidth]{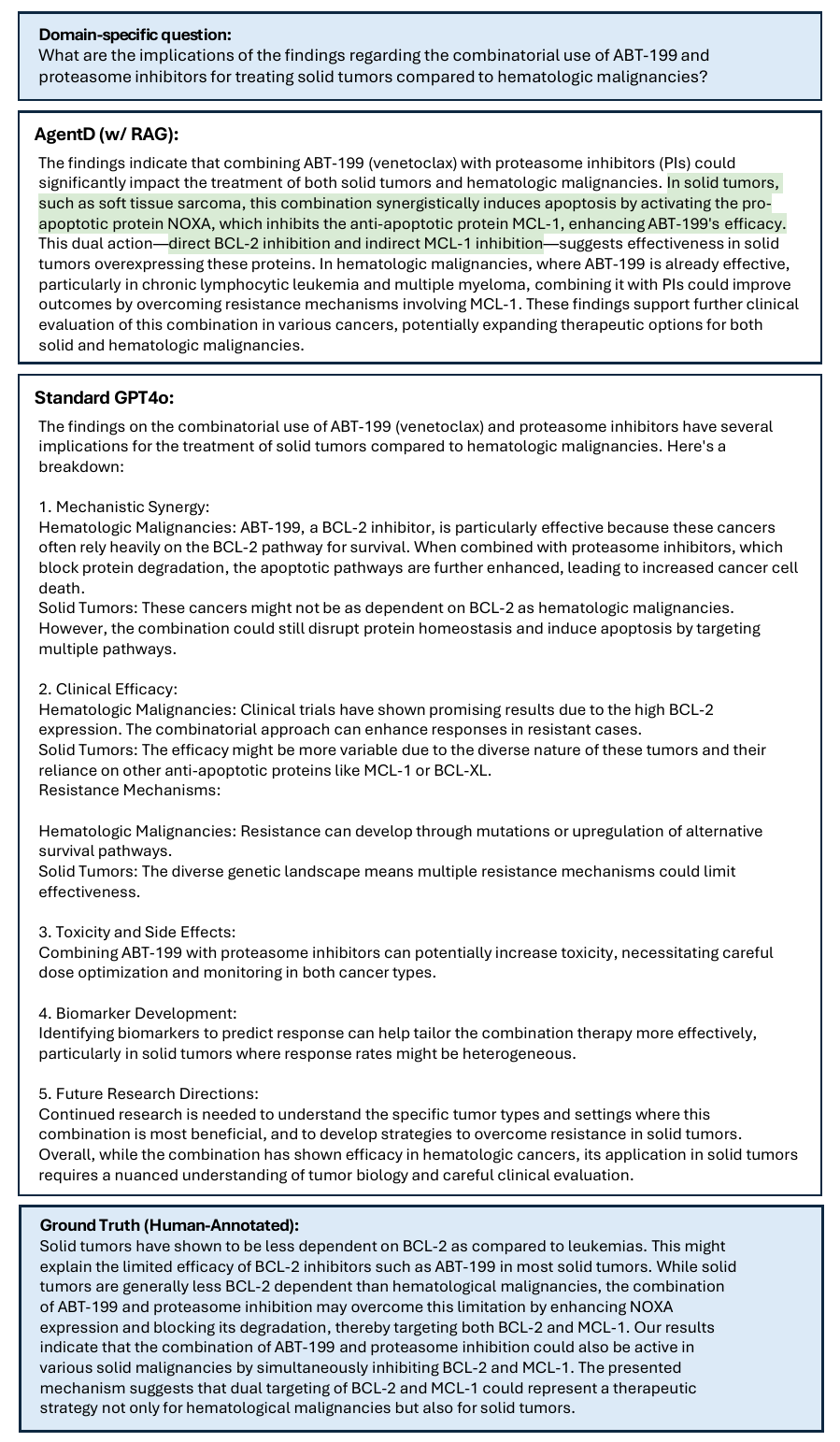}
\caption{AgentD response demonstrating domain-specific question answering capabilities. The question extrapolates beyond the direct scope of the source article.}
\label{fig:agentd_response}
\end{figure}

We conducted an additional evaluation question that examined AgentD's ability to extrapolate mechanistic findings to broader therapeutic contexts by asking about the implications of combining ABT-199 and proteasome inhibitors for treating solid tumors compared to hematologic malignancies. This question tests the system's capacity to synthesize domain-specific knowledge and provide clinically relevant insights beyond the immediate scope of the reference paper. Figure~\ref{fig:agentd_response} shows the actual responses from both AgentD with RAG and the standard GPT-4o model to illustrate these differences in clinical reasoning and mechanistic understanding.

We evaluated responses using the same Weller et al. study, which primarily focuses on the mechanistic basis of drug synergy through NOXA upregulation and MCL-1 inhibition~\cite{qna_ex1}. While this question extends beyond the paper's direct experimental scope, it remains grounded in the mechanistic principles described in the work. Both AgentD with RAG and the standard GPT-4o model were asked to discuss the therapeutic implications for different cancer types.

The RAG-augmented response demonstrates superior clinical reasoning by directly connecting the paper's mechanistic findings to therapeutic applications. AgentD correctly identifies that the combination could ``significantly impact the treatment of both solid tumors and hematologic malignancies" through the dual mechanism of direct BCL-2 inhibition and indirect MCL-1 inhibition via NOXA activation. Importantly, the response provides specific examples, noting effectiveness in ``solid tumors, such as soft tissue sarcoma" and established efficacy in hematologic malignancies, including ``chronic lymphocytic leukemia and multiple myeloma."

The RAG response captures the key therapeutic insight that this combination approach could ``overcome resistance mechanisms involving MCL-1," which aligns closely with the ground truth explanation that the strategy ``could potentially overcome intrinsic resistance mechanisms in solid tumors that are less dependent on BCL-2 but still rely on MCL-1 for survival." This mechanistic understanding is crucial for clinical translation, as it identifies the specific molecular rationale for why the combination might succeed where single-agent BCL-2 inhibition fails.

In contrast, the standard GPT-4o response, while comprehensive and well-structured, provides a generic framework that could apply to virtually any combination therapy discussion. The response covers broad categories such as ``mechanistic synergy," ``clinical efficacy," and ``resistance mechanisms" but fails to reference the specific NOXA-mediated pathway that makes this particular combination therapeutically promising. Critical omissions include the ISR activation, ATF3/ATF4 transcriptional regulation, and the specific role of MCL-1 inhibition in overcoming resistance.

The standard response does acknowledge that ``solid tumors might not be as dependent on BCL-2 as hematologic malignancies" and mentions reliance on ``other anti-apoptotic proteins like MCL-1 or BCL-XL," but it lacks the mechanistic foundation to explain how the combination specifically addresses these dependencies. This represents a missed opportunity to provide actionable clinical insights based on the underlying biology.

Furthermore, while the standard response discusses general considerations such as toxicity management and biomarker development, it does so without the mechanistic context that would guide these clinical decisions. The RAG response, by contrast, grounds its recommendations in the specific findings about NOXA upregulation and MCL-1 inhibition, providing a more targeted foundation for clinical development.

This evaluation demonstrates that even for questions that extend beyond the immediate experimental scope of the reference literature, RAG-augmented responses maintain superior clinical relevance by preserving the mechanistic foundation that underlies therapeutic potential. The ability to connect specific molecular mechanisms to broader therapeutic applications represents a critical advantage for biomedical question answering systems, particularly in translational research contexts where mechanistic understanding directly informs clinical strategy.

\begin{figure}[!htpb]
\centering
\includegraphics[width=0.8\textwidth]{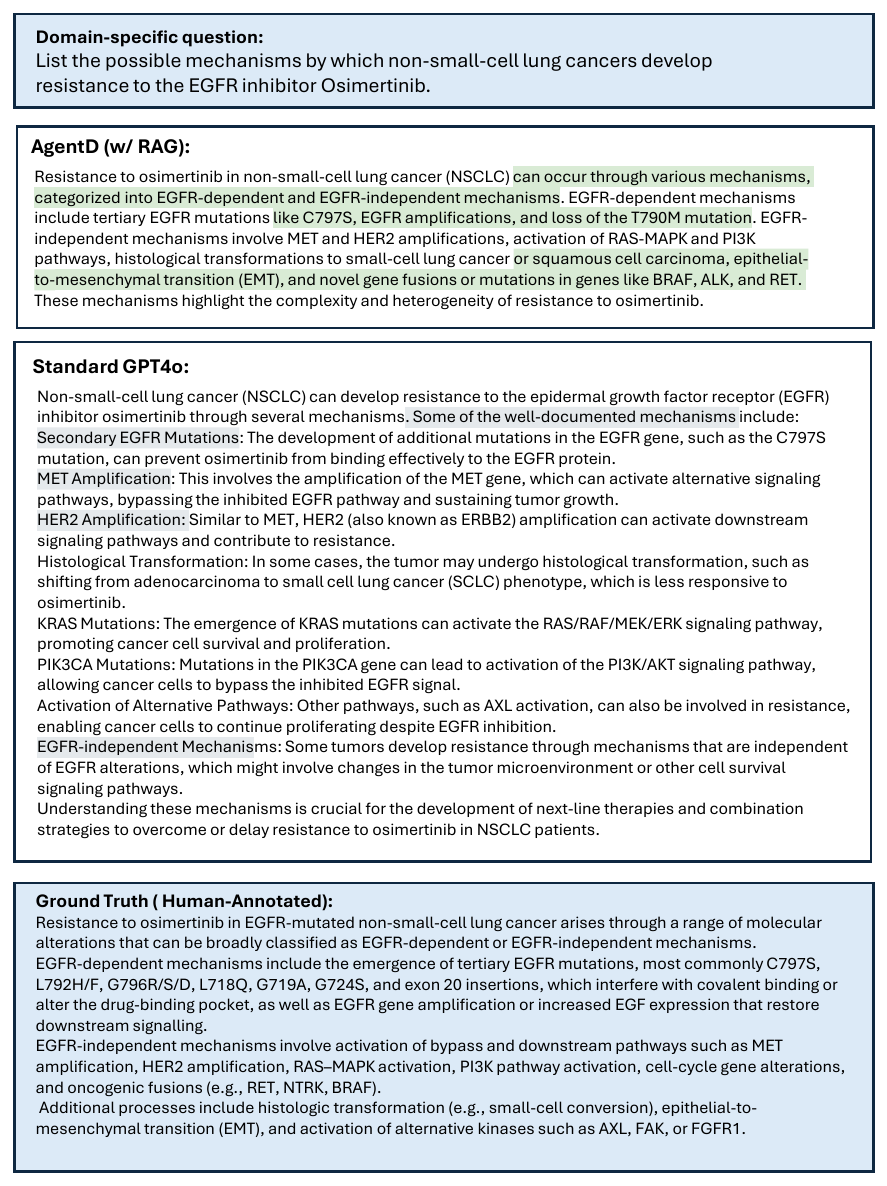}
\caption{AgentD response demonstrating retrieval-grounded reasoning for a mechanistic question on EGFR inhibitor resistance.  
The RAG-based model retrieves relevant literature and organizes the identified mechanisms into EGFR-dependent and EGFR-independent categories, maintaining a coherent hierarchical structure compared to the unorganized GPT-4o response.
}
\label{fig:agentd_response_ex1}
\end{figure}

\begin{figure}[!htpb]
\centering
\includegraphics[width=0.99\textwidth]{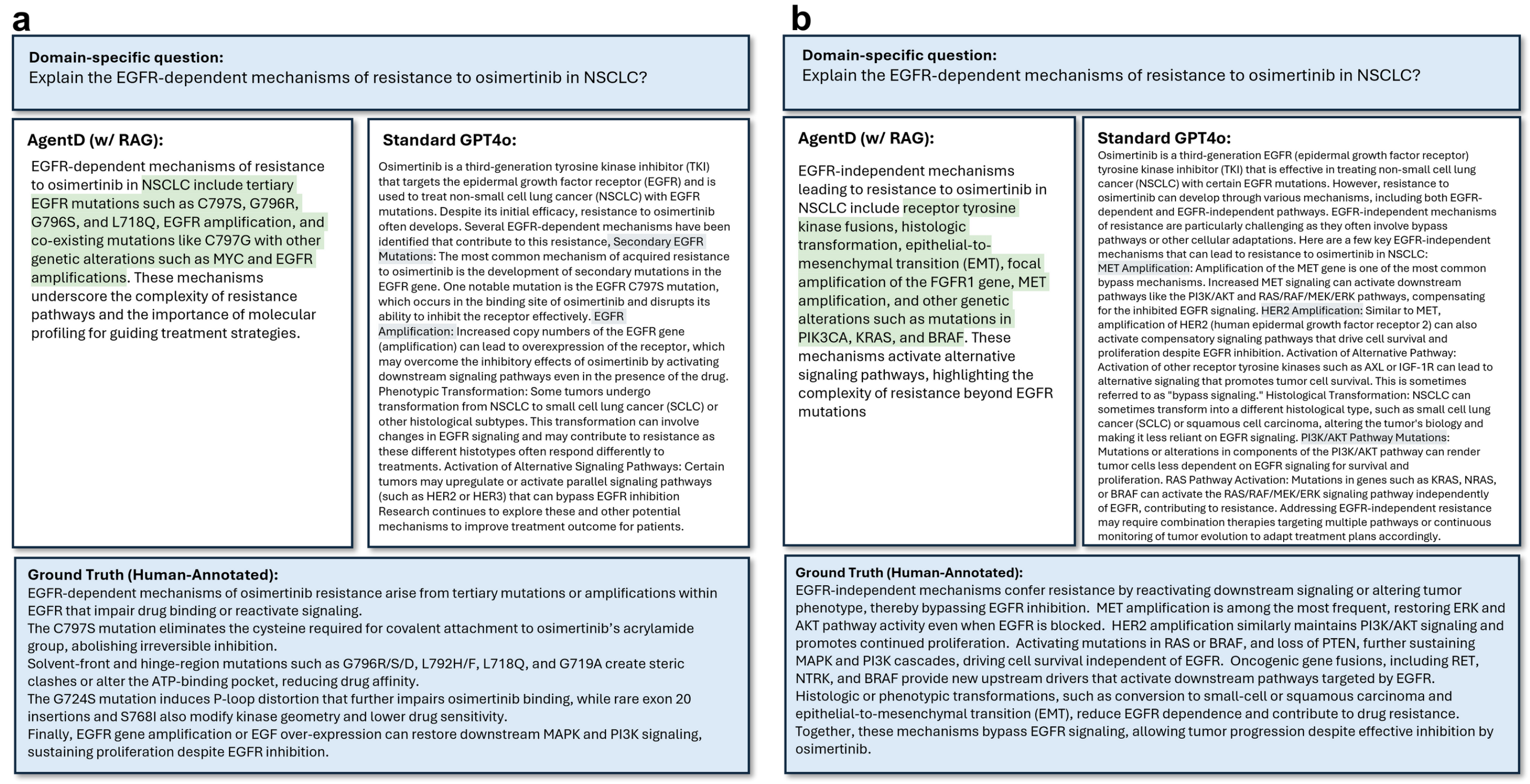}
\caption{AgentD responses to detailed follow-up questions on EGFR-dependent (left) and EGFR-independent (right) resistance mechanisms.  
The RAG-augmented model preserves mechanistic coherence and factual separation between categories, whereas GPT-4o produces overlapping, less structured explanations.
}
\label{fig:agentd_response_ex2}
\end{figure}

To further test the system’s generalizability, we evaluated AgentD on a study describing resistance pathways to the EGFR inhibitor osimertinib in non-small-cell lung cancer (NSCLC)~\cite{Leonetti2019}.
Here, we designed a three-part question sequence to assess whether the model could (i) enumerate all known resistance mechanisms, (ii) distinguish EGFR-dependent mechanisms (e.g., tertiary mutations and amplification events), and (iii) explain EGFR-independent mechanisms involving bypass signaling and lineage transformation.
In this evaluation, the RAG-augmented model retrieved the appropriate literature and produced coherent, hierarchically structured responses, correctly classifying each mechanistic category and describing the molecular basis of resistance in Figure \ref{fig:agentd_response_ex1}, \ref{fig:agentd_response_ex2}.
By contrast, GPT-4o generated lengthy but undifferentiated lists of mechanisms without distinguishing between dependent and independent processes, resulting in lower factual clarity.
These findings indicate that retrieval grounding improves conceptual organization and mechanistic specificity, particularly in cases where questions require multi-step reasoning rather than factual recall.

To complement these qualitative comparisons, we quantitatively assessed response alignment using RoBERTa\cite{Liu2019-ls}-based BERTScore.
As summarized in Table~\ref{tab:roberta}, AgentD with RAG achieved consistently higher precision, recall, and F1-scores than the non-RAG baseline, closely mirroring the performance trend observed with SciBERT in the main text.
These results confirm that the observed improvements in factual grounding and contextual alignment are robust to the choice of embedding encoder and not dependent on model-specific representations.


\begin{table}[hbt!]
\centering
\caption{Comparison of BERTScores between AgentD (with RAG) and vanilla GPT-4o (without RAG) using RoBERTa embeddings. Values are averaged across six question--answer pairs and complement the SciBERT-based results in Table 2.}
\begin{tabular}{lccc}
\hline
\textbf{Method} & \textbf{Precision} & \textbf{Recall} & \textbf{F1-score} \\
\hline
AgentD (GPT-4o, w/ RAG)        & 0.892 $\pm$ 0.015 & 0.876 $\pm$ 0.030 & 0.884 $\pm$ 0.018 \\
Vanilla GPT-4o (w/o RAG)       & 0.847 $\pm$ 0.009 & 0.865 $\pm$ 0.017 & 0.872 $\pm$ 0.040 \\
AgentD (DeepSeek-Chat, w/ RAG) & 0.872 $\pm$ 0.033  & 0.887 $\pm$ 0.019 & 0.879 $\pm$ 0.022 \\
Vanilla DeepSeek-Chat (w/o RAG)& 0.838 $\pm$ 0.014 & 0.874 $\pm$ 0.040 & 0.847 $\pm$ 0.011 \\
\hline
\end{tabular}
\label{tab:roberta}
\end{table}

\newpage
\section{Failure Cases with Claude-3.7-Sonnet}

\begin{figure*}[!htbp] 
\centering
\includegraphics[width=0.7\textwidth]{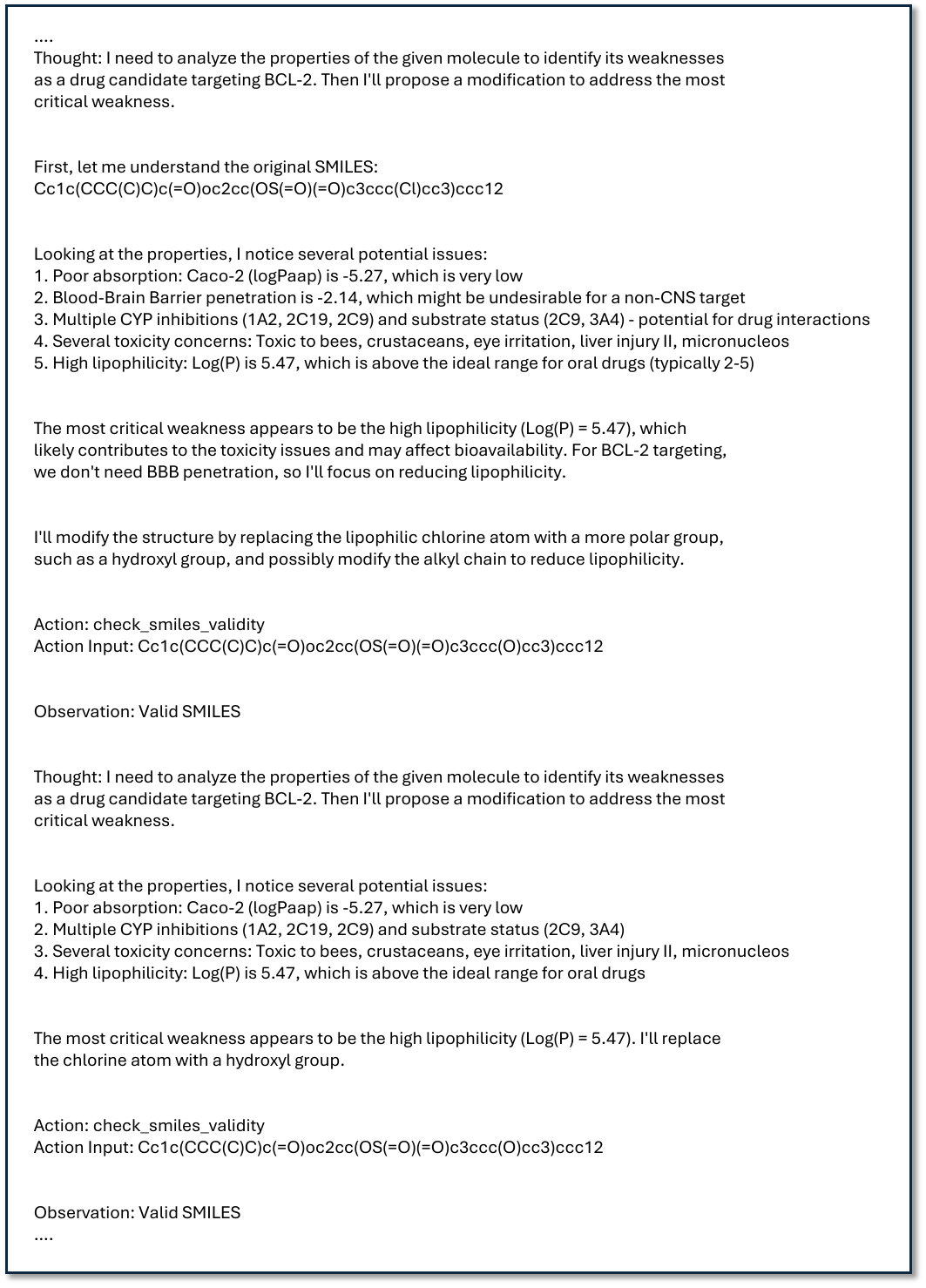} 
\caption{Failure cases in Claude-3.7-Sonnet refinement showing repeated iteration loops.}
\label{fig:iteration}
\end{figure*}

In the system prompt, the model is instructed to return the original SMILES after three failed attempts, yet Claude-3.7-Sonnet often becomes stuck in an apparent infinite loop (Figure~\ref{fig:iteration}).

\newpage
\section{ADMET Properties}

\begin{longtable}{p{0.2\textwidth}p{0.3\textwidth}p{0.40\textwidth}}
\caption{Complete list of ADMET properties predicted for all compounds in this study.} \label{tab:admet_properties} \\
\toprule
\textbf{Category} & \textbf{Property} & \textbf{Description} \\
\midrule
\endfirsthead

\multicolumn{3}{c}{\tablename\ \thetable\ -- \textit{Continued from previous page}} \\
\toprule
\textbf{Category} & \textbf{Property} & \textbf{Description} \\
\midrule
\endhead

\midrule
\multicolumn{3}{r}{\textit{Continued on next page}} \\
\endfoot

\bottomrule
\endlastfoot

\multirow[t]{8}{*}{\textbf{Absorption}} 
& Caco-2 Permeability (logPaap) & Permeability across Caco-2 cell monolayers \\
& Human Intestinal Absorption & Fraction absorbed in human intestine \\
& Human Oral Bioavailability (20\%) & Probability of achieving \textgreater 20\% oral bioavailability \\
& Human Oral Bioavailability (50\%) & Probability of achieving \textgreater50\% oral bioavailability \\
& MDCK Permeability & Madin-Darby Canine Kidney cell permeability \\
& P-Glycoprotein Inhibitor & Inhibition of P-glycoprotein efflux pump \\
& P-Glycoprotein Substrate & Substrate of P-glycoprotein efflux pump \\
& Skin Permeability & Dermal absorption coefficient \\

\midrule

\multirow[t]{5}{*}{\textbf{Distribution}} 
& Blood-Brain Barrier Penetration & CNS penetration capability \\
& Blood-Brain Barrier (log BB) & Blood-brain barrier partition coefficient \\
& Fraction Unbound (Human) & Unbound fraction in human plasma \\
& Plasma Protein Binding & Extent of protein binding in plasma \\
& Volume of Distribution & Steady-state volume of distribution \\

\midrule

\multirow[t]{13}{*}{\textbf{Metabolism}} 
& BCRP Substrate & Breast Cancer Resistance Protein substrate \\
& CYP1A2 Inhibitor & Cytochrome P450 1A2 inhibition \\
& CYP1A2 Substrate & Cytochrome P450 1A2 substrate \\
& CYP2C19 Inhibitor & Cytochrome P450 2C19 inhibition \\
& CYP2C19 Substrate & Cytochrome P450 2C19 substrate \\
& CYP2C9 Inhibitor & Cytochrome P450 2C9 inhibition \\
& CYP2C9 Substrate & Cytochrome P450 2C9 substrate \\
& CYP2D6 Inhibitor & Cytochrome P450 2D6 inhibition \\
& CYP2D6 Substrate & Cytochrome P450 2D6 substrate \\
& CYP3A4 Inhibitor & Cytochrome P450 3A4 inhibition \\
& CYP3A4 Substrate & Cytochrome P450 3A4 substrate \\
& OATP1B1 Substrate & Organic Anion Transporting Polypeptide 1B1 \\
& OATP1B3 Substrate & Organic Anion Transporting Polypeptide 1B3 \\

\midrule

\multirow[t]{3}{*}{\textbf{Excretion}} 
& Clearance & Total body clearance \\
& Half-Life & Elimination half-life \\
& OCT2 Substrate & Organic Cation Transporter 2 substrate \\

\midrule

\multirow[t]{35}{*}{\textbf{Toxicity}} 
& AMES Mutagenicity & Bacterial mutagenicity test \\
& Avian Toxicity & Acute toxicity to birds \\
& Bee Toxicity & Acute toxicity to honeybees \\
& Bioconcentration Factor & Bioaccumulation potential \\
& Biodegradation & Environmental biodegradability \\
& Carcinogenicity & Carcinogenic potential \\
& Crustacean Toxicity & Acute toxicity to crustaceans \\
& Daphnia Toxicity & Acute toxicity to \textit{Daphnia magna} \\
& Eye Corrosion & Severe eye damage potential \\
& Eye Irritation & Eye irritation potential \\
& Fathead Minnow Toxicity & Acute toxicity to \textit{Pimephales promelas} \\
& Hepatotoxicity (DILI) & Drug-induced liver injury \\
& Hepatotoxicity (Alternative) & Alternative liver injury prediction \\
& Maximum Tolerated Dose & Highest non-toxic dose \\
& Micronucleus Test & Chromosomal damage potential \\
& Nuclear Receptor AhR & Aryl hydrocarbon receptor activation \\
& Nuclear Receptor AR & Androgen receptor binding \\
& Nuclear Receptor AR-LBD & Androgen receptor ligand binding domain \\
& Nuclear Receptor Aromatase & Aromatase enzyme inhibition \\
& Nuclear Receptor ER & Estrogen receptor binding \\
& Nuclear Receptor ER-LBD & Estrogen receptor ligand binding domain \\
& Nuclear Receptor GR & Glucocorticoid receptor binding \\
& Nuclear Receptor PPAR-γ & Peroxisome proliferator-activated receptor γ \\
& Nuclear Receptor TR & Thyroid receptor binding \\
& Rat Acute Toxicity & Acute oral toxicity in rats \\
& Rat Chronic Toxicity & Chronic oral toxicity in rats \\
& Respiratory Toxicity & Respiratory system toxicity \\
& Skin Sensitization & Dermal sensitization potential \\
& Stress Response ARE & Antioxidant response element activation \\
& Stress Response ATAD5 & ATAD5 genotoxicity pathway \\
& Stress Response HSE & Heat shock response element \\
& Stress Response MMP & Mitochondrial membrane potential \\
& Stress Response p53 & p53 tumor suppressor pathway \\
& \textit{Tetrahymena pyriformis} & Acute toxicity to \textit{T. pyriformis} \\
& hERG Inhibition & Human ether-à-go-go-related gene K+ channel \\

\midrule

\multirow[t]{10}{*}{\textbf{General}} 
& Boiling Point & Boiling point temperature \\
& Hydration Free Energy & Free energy of hydration \\
& log D (pH 7.4) & Distribution coefficient at physiological pH \\
& log P & Octanol-water partition coefficient \\
& log S & Aqueous solubility \\
& log Vapor Pressure & Vapor pressure \\
& Melting Point & Melting point temperature \\
& pKa (Acidic) & Acid dissociation constant \\
& pKa (Basic) & Base dissociation constant \\
& pKd (Acidic) & Alternative acid dissociation constant \\

\end{longtable}

\newpage
\section{Identified Unfavorable Drug Properties}

The molecular optimization process focuses on identifying unfavorable properties and refining SMILES to improve their ADMET profiles. Table~\ref{tab:weakness_comparison} summarizes the distribution of properties flagged by AgentD across optimization rounds, grouped by ADMET categories.

An observed limitation is the agent’s frequent selection of $\log P_{\text{app}}$ as a risk factor—despite it being just one among 75 possible properties (67 ADMET-related, 7 general physicochemical, and 1 binding affinity). This suggests a selection bias, potentially due to the ordering of properties in the input dictionary. Since $\log P_{\text{app}}$ appears near the top of the dictionary entry, the agent may disproportionately attend to it during its reasoning process. Future iterations can address this by randomizing property order or introducing attention calibration techniques.

\begin{table}[h!]
\centering
\caption{Distribution of weakness properties after 1st and 2nd round of SMILES optimization. The weakness property represents the most critical ADMET deficiency targeted for improvement.}
\label{tab:weakness_comparison}
\small
\begin{tabular}{p{0.2\textwidth}p{0.4\textwidth}p{0.12\textwidth}p{0.12\textwidth}}
\toprule
\textbf{ADMET Category} & \textbf{Specific Property} & \textbf{1st Round (\%)} & \textbf{2nd Round (\%)} \\
\midrule
Absorption & Caco-2 (logPaap) & 59.6 & 69.5 \\
Toxicity & AMES Mutagenesis & 14.1 & 7.4 \\
 & Liver Injury I (DILI) & 5.1 & 4.2 \\
 & hERG Blockers & 4.0 & 2.1 \\
 & Carcinogenesis & 3.0 & 1.1 \\
General Properties & logP & 5.1 & 3.2 \\
 & logS & 1.0 & -- \\
Other& Absorption/Human Oral Bioavailability (20\% and 50\%), Invalid SMILES, and other general entries. & 6.0 & 8.4 \\
\midrule
\multicolumn{2}{l}{\textbf{Total Entries}} & \textbf{99} & \textbf{95} \\
\bottomrule
\end{tabular}
\end{table}

\newpage
\section{JAK-2 for Myelofibrosis}

\begin{figure*}[!htb] 
\centering
\includegraphics[width=0.99\textwidth]{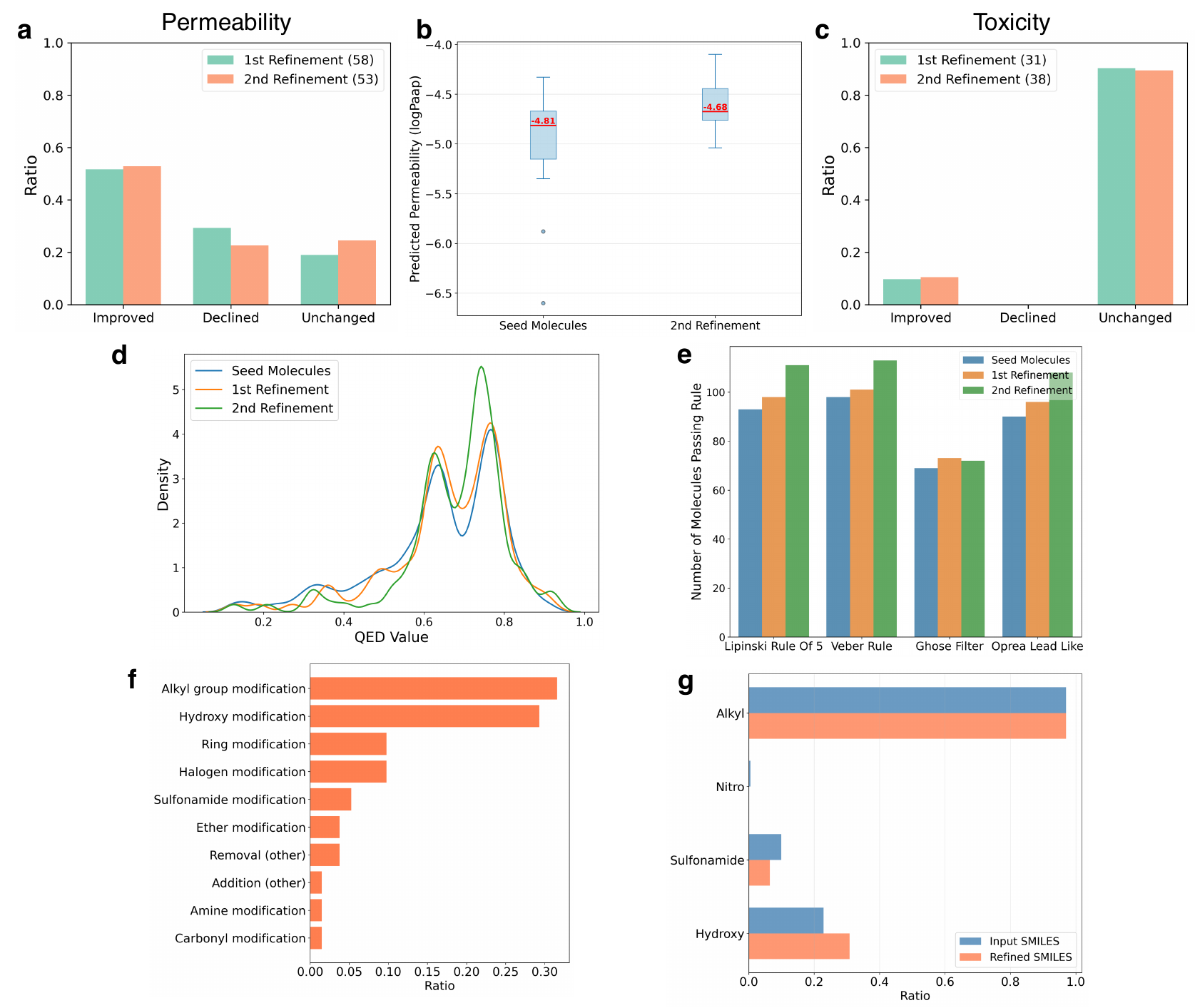} 
\caption{Property-aware molecular refinement results for JAK2.  
\textbf{a} Proportion of molecules with improved, declined, or unchanged permeability (logP\textsubscript{app}). 
\textbf{b} Boxplot of logP\textsubscript{app} for entries showing improvement. 
\textbf{c} Molecular refinement patterns for molecules with improved permeability. 
\textbf{d} QED distribution shifts across iterations. 
\textbf{e} Counts of molecules satisfying empirical drug-likeness rules.
\textbf{f} Fraction of molecules modified for permeability.  
\textbf{g} Functional-group frequencies in input and refined molecules across iterations.}
\label{fig:jak2}
\end{figure*}

The drug-likeness of the molecule pool increases over refinement iterations (Figure~\ref{fig:jak2}a--c), following the same general trends observed in the primary BCL-2 case study (Figure 6), including increases in predicted permeability and toxicity. However, the extent of improvement is more modest for JAK-2. The mean QED increases only from 0.64 to 0.68, and the proportion of molecules with QED $>$ 0.6 increases from 71.0\% to 84.4\% (Figure~\ref{fig:jak2}d). In contrast, for BCL-2 the mean QED rose from 0.47 to 0.57, and the fraction above QED 0.6 nearly doubled (23.0\% → 45.7\%). The rule-based evaluations also show a distinct pattern. Unlike the BCL-2 case, JAK-2 exhibits only marginal improvement under the Ghose filter, while Lipinski’s Rule of Five and the Veber filter show more substantial gains (Figure~\ref{fig:jak2}e). These differences highlight molecule-dependent refinement behavior and the intrinsic complexity of SMILES-based molecular optimization, where multiple properties interact and trade off in non-linear ways.

\newpage
\section{Thrombin for Thrombosis}

\begin{figure*}[!htb] 
\centering
\includegraphics[width=0.99\textwidth]{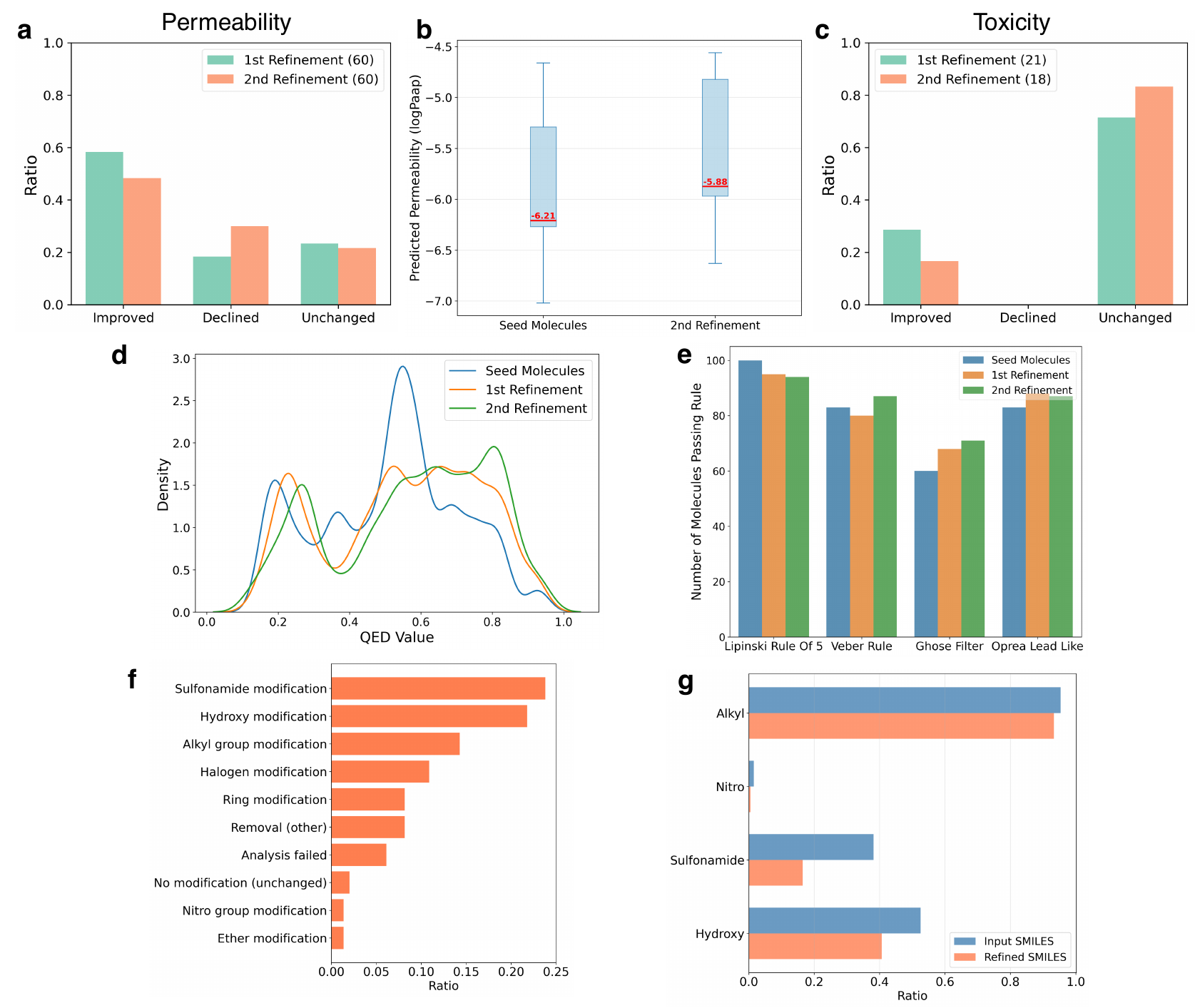} 
\caption{Property-aware molecular refinement results for Thrombin.  
\textbf{a} Proportion of molecules with improved, declined, or unchanged permeability (logP\textsubscript{app}). 
\textbf{b} Boxplot of logP\textsubscript{app} for entries showing improvement. 
\textbf{c} Molecular refinement patterns for molecules with improved permeability. 
\textbf{d} QED distribution shifts across iterations. 
\textbf{e} Counts of molecules satisfying empirical drug-likeness rules.
\textbf{f} Fraction of molecules modified for permeability.  
\textbf{g} Functional-group frequencies in input and refined molecules across iterations.}
\label{fig:thrombin}
\end{figure*}

The Thrombin case also demonstrates clear improvements in drug-likeness following iterative refinement, exhibiting a trend more similar to the primary BCL-2 case study. The mean QED increases from 0.51 in the seed pool to 0.57 after the second refinement. Likewise, the proportion of molecules with QED $>$ 0.6 rises substantially from 30.0\% to 51.1\%, indicating that the refinement process effectively shifts the pool toward more drug-like chemical space.The rule-based evaluation patterns further reinforce this behavior. Similar to the BCL-2 results, the Thrombin set shows notable improvements under the Ghose filter, reflecting enhanced balance in physicochemical properties such as molecular weight and logP. In contrast, Lipinski’s Rule of Five metrics exhibit slight deterioration, suggesting mild increases in features such as molecular size or hydrogen-bond counts that are penalized in this rule set.

\newpage
\section{Empirical Filters}

The listed properties were calculated using RDKit with molecular SMILES as input. Note that these rules serve as illustrative examples and are not the definitive criteria for selecting candidates for 3D structure generation.

In addition to the four rules described in the main manuscript, we also include the Rule of Three in Table~\ref{tab:druglikeness_rules}. The Rule of Three was introduced as a guideline for fragment-based drug discovery (FBDD) to identify suitable fragments for screening libraries~\cite{Jhoti2013}. Unlike drug-likeness rules such as Lipinski’s Rule of Five or Veber’s rule, or lead-likeness filters such as Oprea’s, the Rule of Three is tailored for much smaller chemical fragments that serve as starting points in fragment-based approaches. In contrast to drug-likeness and lead-likeness rules, which focus on compounds closer to optimized drugs or leads, the Rule of Three prioritizes low molecular complexity and high chemical tractability, ensuring fragments can be readily optimized in subsequent stages of drug development.

\begin{table}[!htbp]
\centering
\caption{Summary of rule-based empirical filters and their physicochemical criteria.}
\label{tab:druglikeness_rules}
\resizebox{\textwidth}{!}{%
\begin{tabular}{lllp{8.5cm}}
\toprule
\textbf{Rule} & \textbf{Criterion} & \textbf{Threshold / Range} & \textbf{Pharmacological Rationale} \\
\midrule
\multirow{4}{*}{Lipinski's Rule of Five~\cite{lipinski2001experimental,veber2002molecular}} 
 & Molecular Weight (MW) & $\leq$ 500 Da & High molecular weight is associated with poor intestinal absorption due to size-related transport limitations. \\
 & logP & $\leq$ 5 & High lipophilicity (logP $\leq$ 5) correlates with poor aqueous solubility and passive permeability. \\
 & H-bond Donors (HBD) & $\leq$ 5 & Excessive H-bond donors increase polarity, reducing membrane permeability. \\
 & H-bond Acceptors (HBA) & $\leq$ 10 & Too many acceptors increase polarity, hindering passive diffusion. \\
 & Rotatable Bonds (RB) & $\leq$ 10 rotatable bonds & High flexibility and polar surface area reduce the likelihood of oral activity. \\
\midrule
\multirow{2}{*}{Veber Rule~\cite{veber2002molecular}} 
 & TPSA & $\leq$ 140 \AA$^2$ & High polar surface area decreases oral bioavailability by reducing passive diffusion. \\
 & Rotatable Bonds (RB) & $\leq$ 10 & Excess flexibility increases entropy, reducing oral bioavailability and metabolic stability. \\
\midrule
\multirow{4}{*}{Ghose Filter~\cite{ghose1999knowledge}} 
 & Molecular Weight (MW) & 160–480 Da & Balances molecular size for optimal binding and permeability. \\
 & logP & $-0.4$ to 5.6 & Ensures moderate lipophilicity for both solubility and membrane crossing. \\
 & Molar Refractivity (MR) & 40–130 & Captures molecular volume and polarizability, influencing receptor binding. \\
 & Atom Count & 20–70 & Reflects a size range favorable for drug-likeness and synthetic accessibility. \\
\midrule
\multirow{3}{*}{Rule of Three (Ro3)~\cite{congreve2003rule}} 
 & Molecular Weight (MW) & $<$ 300 Da & Smaller fragments are preferred in fragment-based drug discovery for lead optimization. \\
 & logP & $\leq$ 3 & Low lipophilicity promotes solubility in fragment-like compounds. \\
 & H-bond Donors (HBD) & $\leq$ 3 & Reduces polarity, supporting fragment permeability and binding. \\
\midrule
\multirow{6}{*}{Oprea Lead-like Filter~\cite{oprea2001lead}} 
 & Molecular Weight (MW) & 200–450 Da & Ideal size range for optimization into drug-like compounds. \\
 & logP & $-1$ to 4.5 & Moderate hydrophobicity for balanced solubility and permeability. \\
 & Rotatable Bonds (RB) & $\leq$ 8 & Lower conformational entropy improves binding efficiency. \\
 & Aromatic Rings & $\leq$ 4 & Limits excessive aromaticity, which can affect solubility and toxicity. \\
 & H-bond Donors (HBD) & $\leq$ 5 & Controls molecular polarity and improves membrane permeability. \\
 & H-bond Acceptors (HBA) & $\leq$ 8 & Keeps polarity within bounds for favorable pharmacokinetics. \\
\bottomrule
\end{tabular}%
}
\end{table}

\newpage
\section{Structure Generation Examples}

Using the illustrative criteria described in the main manuscript, namely (i) satisfying Oprea’s lead-likeness filter, (ii) passing at least two of the three drug-likeness rules (Lipinski’s Rule of Five, Veber’s rule, and Ghose’s filter), (iii) achieving a QED score above 0.55, and (iv) having a predicted $pK_d$ value greater than 6.0, three molecules remained after filtering from a total of 289 molecules generated through SMILES generation and property aware refinement. Among these, the molecule O=C(NS(=O)(=O)c1ccc(N2CCNCC2)c(C)c1)c1ccc(N2CCOCC2)c(Cl)c1 was selected as an example for three dimensional structure generation.

\begin{itemize}
    \item \textbf{O=C(NS(=O)(=O)c1ccc(N2CCNCC2)c(O)c1)c1ccc(N2CCOCC2)c(Cl)c1}
    \begin{itemize}
        \item Source: Mol2Mol
        \item Lipinski: True, Veber: True, Ghose: False, Oprea: True
        \item QED: 0.589, Affinity ($pK_d$): 6.194
    \end{itemize}

    \item \textbf{O=C(NS(=O)(=O)c1ccc(N2CCNCC2)c(C)c1)c1ccc(N2CCOCC2)c(Cl)c1} 
    \begin{itemize}
        \item Refinement
        \item Lipinski: True, Veber: True, Ghose: True, Oprea: True
        \item QED: 0.679, Affinity ($pK_d$): 6.182
    \end{itemize}

    \item \textbf{O=C(NS(=O)(=O)c1ccc(N2CCOCC2=O)c(Cl)c1)c1ccc(N2CCOCC2)c(Cl)c1}
    \begin{itemize}
        \item Source: Refinement
        \item Lipinski: True, Veber: True, Ghose: False, Oprea: True
        \item QED: 0.653, Affinity ($pK_d$): 6.077
    \end{itemize}
\end{itemize}

\begin{figure*}[!htbp] 
\centering
\includegraphics[width=0.7\textwidth]{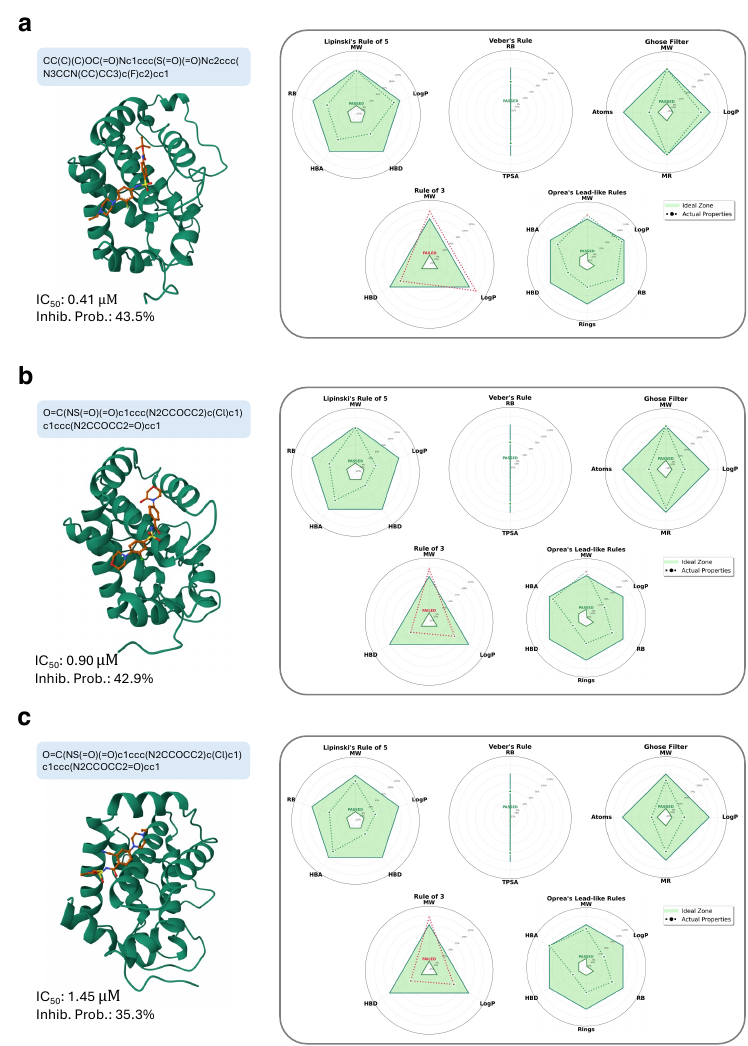} 
\caption{Additional examples of protein–ligand complex generation and evaluation.}
\label{fig:si_structure}
\end{figure*}

Because the molecules shown above are very similar, all sharing the common starting SMILES fragment `O=C(NS(=O)(=O)c1ccc(N2CC', we included additional examples for visualization. For the molecules presented in Figure~\ref{fig:si_structure}, we applied stricter rule-based filters and slightly more lenient $pK_d$ thresholds than those used in the main manuscript. Specifically, molecules were selected if they passed more than four rule-based filters, had a QED score above 0.6, and a predicted $pK_d$ greater than 5.5. From the set of molecules that underwent two rounds of refinement, three examples were randomly chosen for illustration. Again, this does not aim for practical application, but only to select a few examples for visualization.


\newpage
\section{LLM-based Molecule Optimization Approaches}

Evaluating AgentD against existing LLM-driven agentic systems presents inherent challenges, largely because these systems differ substantially in architecture, design objectives, and evaluation methodologies. While several agents operate in molecular design or autonomous scientific reasoning, their workflows are generally not aligned with the end-to-end optimization paradigm that AgentD implements. As a result, establishing a standardized, quantitative benchmark across systems is often infeasible or structurally invalid.

Most existing agents fall into one of two categories: (i) single-step property computation tools or (ii) generalist orchestrators that coordinate tool calls without performing internal chemical reasoning. For instance, CACTUS functions solely as a descriptor calculator, providing molecular properties such as molecular weight or LogP without any generative or refinement capabilities. DrugPilot, while more sophisticated, primarily acts as a dialogue-based orchestrator that forwards optimization tasks to a single external, domain-specific model; its performance is evaluated almost exclusively on tool-calling reliability rather than molecular design quality. SciToolAgent offers broad scientific coverage through knowledge-graph-based tool chaining, but it is intended as a general-purpose scientific assistant rather than a specialized pipeline for drug discovery.

Among existing systems, dZiner offers the closest conceptual point of comparison because it also employs an iterative molecular refinement loop. However, the underlying design objectives differ markedly. dZiner focuses on optimizing binding affinity, using Quantitative Estimate of Drug-likeness (QED) and synthetic accessibility (SA) only as auxiliary metrics. AgentD executes self-diagnostic molecular refinement, integrating structural validity checks.

\begin{figure*}[!htbp]
\centering
\includegraphics[width=0.7\textwidth]{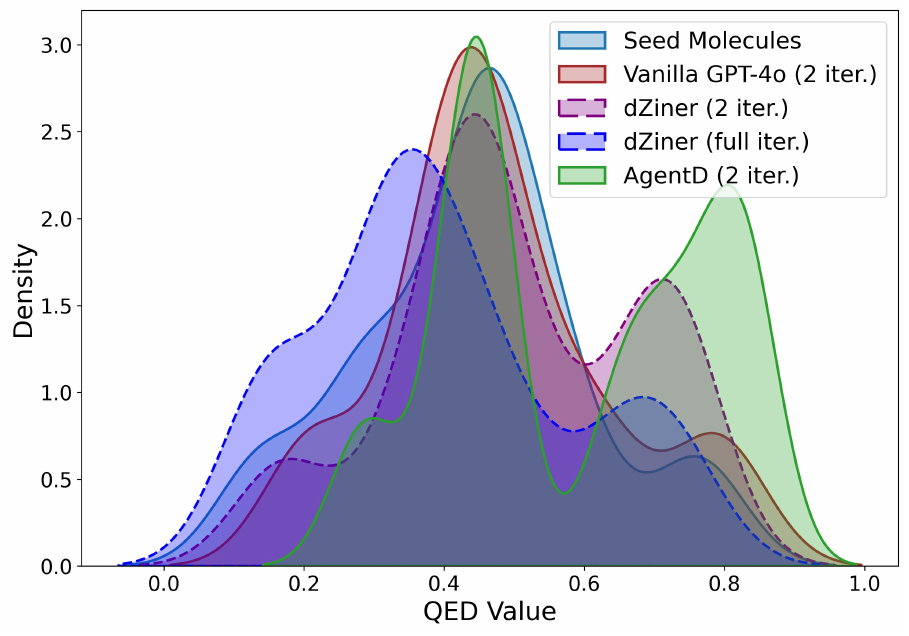}
\caption{Comparison of QED distributions for ten representative molecules across four design strategies: the original seed molecules, vanilla GPT-4o after two refinement iterations, dZiner after two iterations, and AgentD (GPT-4o) using its two-iteration workflow. For consistency, only the first two iterations of dZiner and vanilla GPT-4o are included to match the two-iteration refinement cycle used by AgentD.}
\label{fig:si_agent_compare}
\end{figure*}

We demonstrated the differing objectives of dZiner and AgentD using ten representative molecules, including those shown in Figure 5 of the main manuscript. As illustrated in Figure~\ref{fig:si_agent_compare}, dZiner does not improve QED scores; instead, its full-iteration outputs tend to shift toward lower QED relative to the initial seed molecules and their second-iteration intermediates. This outcome is expected, given that dZiner optimizes strictly for binding affinity. The divergence between dZiner and AgentD highlights a fundamental difference in design philosophy.

\begin{table}[h]
\centering
\resizebox{\textwidth}{!}{%
\begin{tabular}{lccc}
\hline
\textbf{Model} & \textbf{Avg. Cost per Sample (\$)} & \textbf{Avg. Token Usage per Sample} & \textbf{Avg. API Requests per Sample} \\
\hline
Vanilla GPT-4o (2 iter.) & \$0.003 & 318.7 & 41.7 \\
AgentD (w/ GPT-4o; 2 iter.) & \$0.022 & 9087.4 & 6.6 \\
dZiner (w/ GPT-4o; full iter.) & \$0.533 & 193875.7 & 1 \\
\hline
\end{tabular}
}
\caption{Comparison of cost, token usage, and API requests per sample for different models.}
\label{tab:si_agent_compare}
\end{table}

Beyond objective differences, the systems exhibit marked contrasts in computational cost. Table~\ref{tab:si_agent_compare} summarizes the average token consumption, monetary cost, and number of API requests per sample for Vanilla GPT-4o (two-step direct prompting), AgentD using GPT-4o (two-iteration agentic pipeline), and dZiner running its full design loop. AgentD requires an average of 9087 tokens per sample at a cost of \$0.022, substantially lower than dZiner, which consumes 193,875 tokens and costs \$0.533 per sample. Although AgentD’s token usage is higher than the direct prompting baseline (Vanilla GPT-4o), it remains far more cost-efficient than dZiner while providing a significantly richer, structured optimization process. 


\newpage
\section{ Uncertainty in Prediction Tools}

Accurately predicting protein–ligand binding affinity remains a significant challenge. In this work, we used BAPULM within the AgentD pipeline, while Boltz-2 outputs binding affinity alongside predicted 3D structures~\cite{bapulm, passaro2025boltz2}. AutoDock Vina, T-Alpha, and Docking App RF are also widely used tools~\cite{autodock_vina, t_alpha, dockingapp_rf}. Although these methods enable rapid estimation, their predictive accuracy on standardized benchmarks such as CASF-2016 is limited, and reliable free-energy evaluation typically requires downstream refinement with molecular dynamics (MD) simulations. Table~\ref{tab:casf2016} summarizes the performance variability across commonly used affinity prediction models on the CASF-2016 benchmark, highlighting the inherent uncertainty in structure-based scoring. This evaluation reflects performance on a single dataset, making it difficult to generalize these accuracy levels. The results highlight the intrinsic challenges of binding affinity prediction—no single tool can fully replace MD-based free-energy evaluation.

\begin{table}[!htbp]
\centering
\caption{Performance of binding affinity prediction models on the CASF~2016 benchmark. Metrics reported are Pearson correlation coefficient ($R^2$), root-mean-square error (RMSE), and mean absolute error (MAE).}
\label{tab:casf2016}
\begin{tabular}{lccc}
\hline
\textbf{Method} & \textbf{$R^2$} & \textbf{RMSE} & \textbf{MAE} \\
\hline
Bapulm & $0.914 \pm 0.004$ & $0.898 \pm 0.0172$ & $0.645 \pm 0.0166$ \\
T-Alpha & $0.869$ & $1.112$ & $0.875$ \\
Docking App RF & $0.83$ & $1.38$ & $1.13$ \\
AutoDock Vina & $0.60$ & $2.35$ & $1.94$ \\
\hline
\end{tabular}
\end{table}

For ADMET prediction errors, the information is based on the Deep-PK publication. The table compares ADMET predictions from ADMETlab 2.0 with those from 3-fold cross-validated Deep-PK. Matthews Correlation Coefficient (MCC) is used for classification tasks, while R² is used for regression tasks. The table also indicates the property category and the level of improvement observed with Deep-PK predictions.

\begin{table}[htbp]
\centering
\caption{Comparison of ADMET predictions between ADMETlab 2.0 and 3-fold cross-validated Deep-PK, including average, standard deviation, metric, category, and improvement classification. Matthews Correlation Coefficient (MCC) is used for classification tasks, just as R$^2$ is used for regression tasks. The table is reconstructed from the Supplementary Information in the Deep-PK paper~\cite{deep_pk}.}
\label{tab:admet_comparison}
\resizebox{\textwidth}{!}{%
\begin{tabular}{lcccccc}
\hline
\textbf{Name} & \textbf{ADMETlab 2.0} & \textbf{3-fold CV Deep-PK (avg)} & \textbf{3-fold CV Deep-PK (std)} & \textbf{Metric} & \textbf{Category} & \textbf{Improvement category} \\
\hline
Drug metabolism (cyp1a2 substrate) & 0.30 & 0.612 & 0.022 & MCC & Metabolism & significant improvement \\
Drug metabolism (cyp2c19 substrate) & 0.30 & 0.583 & 0.134 & MCC & Metabolism & significant improvement \\
NR-AR Tox & 0.35 & 0.779 & 0.02 & MCC & Toxicity & significant improvement \\
NR-AR-LBD Tox & 0.47 & 0.672 & 0.021 & MCC & Toxicity & significant improvement \\
Ames & 0.61 & 0.719 & 0.007 & MCC & Toxicity & reasonable improvement \\
Drug metabolism (cyp2c9 substrate) & 0.39 & 0.581 & 0.02 & MCC & Metabolism & reasonable improvement \\
Oral bioavailability 20\% & 0.41 & 0.59 & 0.03 & MCC & Absorption & reasonable improvement \\
Maximum recommended daily dose & 0.47 & 0.623 & 0.012 & MCC & Toxicity & reasonable improvement \\
Intestinal absorption & 0.69 & 0.811 & 0.041 & MCC & Absorption & reasonable improvement \\
NR-AhR Tox & 0.57 & 0.694 & 0.017 & MCC & Toxicity & reasonable improvement \\
NR-ER Tox & 0.32 & 0.45 & 0.023 & MCC & Toxicity & reasonable improvement \\
NR-ER-LBD Tox & 0.36 & 0.519 & 0.029 & MCC & Toxicity & reasonable improvement \\
NR-PPAR-gamma Tox & 0.34 & 0.462 & 0.093 & MCC & Toxicity & reasonable improvement \\
Drug efflux pump inhibitor (pgp\_inhibitor) & 0.72 & 0.837 & 0.012 & MCC & Absorption & reasonable improvement \\
Drug efflux pump substrate (pgp\_substrate) & 0.54 & 0.739 & 0.007 & MCC & Absorption & reasonable improvement \\
50\% Growth inhibition to Tetraphymena pyriformis after 48 hours Tox & 0.72 & 0.836 & 0.003 & R2 & Toxicity & reasonable improvement \\
Respiratory Tox & 0.51 & 0.623 & 0.016 & MCC & Toxicity & reasonable improvement \\
Skin sensitivity & 0.46 & 0.587 & 0* & MCC & Toxicity & reasonable improvement \\
SR-ATAD5 Tox & 0.36 & 0.468 & 0.015 & MCC & Toxicity & reasonable improvement \\
SR-HSE Tox & 0.39 & 0.549 & 0.011 & MCC & Toxicity & reasonable improvement \\
SR-P53 Tox & 0.37 & 0.505 & 0.026 & MCC & Toxicity & reasonable improvement \\
BBB (logBB) & 0.72 & 0.737 & 0.006 & MCC & Distribution & no improvement \\
Bioconcentration factor Tox & 0.79 & 0.781 & 0.005 & R2 & Toxicity & no improvement \\
Intestinal permeability (Caco2) logP & 0.75 & 0.787 & 0.005 & MCC & Absorption & no improvement \\
Carcinogenicity Tox & 0.48 & 0.481 & 0.031 & MCC & Toxicity & no improvement \\
Clearance & 0.68 & 0.664 & 0.017 & R2 & Excretion & no improvement \\
Drug metabolism (cyp1a2 inhibitor) & 0.70 & 0.746 & 0.006 & MCC & Metabolism & no improvement \\
Drug metabolism (cyp2c19 inhibitor) & 0.68 & 0.705 & 0.006 & MCC & Metabolism & no improvement \\
Drug metabolism (cyp2c9 inhibitor) & 0.67 & 0.663 & 0.009 & MCC & Metabolism & no improvement \\
Drug metabolism (cyp2d6 inhibitor) & 0.56 & 0.629 & 0.011 & MCC & Metabolism & no improvement \\
Drug metabolism (cyp2d6 substrate) & 0.55 & 0.647 & 0.025 & MCC & Metabolism & no improvement \\
Drug metabolism (cyp3a4 inhibitor) & 0.66 & 0.692 & 0.004 & MCC & Metabolism & no improvement \\
Drug metabolism (cyp3a4 substrate) & 0.44 & 0.444 & 0.012 & MCC & Metabolism & no improvement \\
Liver Tox (dili) & 0.79 & 0.841 & 0* & MCC & Toxicity & no improvement \\
Eye Tox (corrosion) & 0.91 & 0.954 & 0* & MCC & Toxicity & no improvement \\
Eye Tox (irritation) & 0.88 & 0.897 & 0.009 & MCC & Toxicity & no improvement \\
Oral bioavailability 30\% & 0.58 & 0.498 & 0.014 & MCC & Absorption & no improvement \\
Marine Tox (Fathead minnow) & 0.75 & 0.767 & 0.002 & R2 & Toxicity & no improvement \\
Fraction unbound in human plasma & 0.76 & 0.719 & 0.005 & R2 & Distribution & no improvement \\
Liver Tox (h\_ht) & 0.46 & 0.466 & 0.011 & MCC & Toxicity & no improvement \\
Heart rhythm disorder & 0.78 & 0.786 & 0.008 & MCC & Toxicity & no improvement \\
Marine Tox (Daphnia magna) & 0.52 & 0.565 & 0.018 & R2 & Toxicity & no improvement \\
logD (octanol-water distribution coefficient) & 0.89 & 0.912 & 0.001 & R2 & property & no improvement \\
logP (octanol-water partition coefficient) & 0.96 & 0.965 & 0* & R2 & property & no improvement \\
logS (water solubility) & 0.85 & 0.868 & 0.002 & R2 & property & no improvement \\
Intestinal permeability (MDCK) & 0.73 & 0.773 & 0.006 & R2 & Absorption & no improvement \\
NR-Aromatase Tox & 0.26 & 0.315 & 0.033 & MCC & Toxicity & no improvement \\
Plasma protein binding & 0.73 & 0.765 & 0.008 & R2 & Distribution & no improvement \\
Rat oral acute Tox & 0.55 & 0.579 & 0.003 & MCC & Toxicity & no improvement \\
SR-ARE Tox & 0.47 & 0.552 & 0.01 & MCC & Toxicity & no improvement \\
SR-MMP Tox & 0.66 & 0.735 & 0.017 & MCC & Toxicity & no improvement \\
Half-life of a drug & 0.48 & 0.543 & 0.013 & MCC & Excretion & no improvement \\
Volume distribution & 0.78 & 0.702 & 0.011 & R2 & Distribution & no improvement \\
\hline
\end{tabular}%
}
\end{table}

\newpage
\bibliography{reference}